\documentclass[]{article}
\usepackage{arxiv}
\usepackage{tikz}
\usepackage{soul}
\usepackage{xcolor}
\usepackage{mathtools}
\usepackage{adjustbox}
\usepackage{tabularx}
\usepackage{adjustbox}
\usepackage{siunitx}
\usepackage{subcaption}
\usepackage{bm}
\usepackage{upgreek}
\usepackage[square,sort,comma,authoryear,round]{natbib}
\usepackage{hyperref}
\hypersetup{colorlinks,allcolors=black}
\usepackage{multirow}
\usepackage{makecell}
\usepackage{graphicx}
\usepackage{amssymb}
\usepackage{amsthm}
\usepackage{booktabs}
\usepackage{amsmath}
\usepackage[utf8]{inputenc}
\usepackage{varioref}
\usepackage{cleveref}
\usepackage{graphicx}
\usepackage{enumitem}
\usepackage{amsmath}
\usepackage{multirow}
\usepackage{tcolorbox}
\usepackage{xcolor}
\usepackage{adjustbox}
\usepackage{appendix}
\usepackage{url}
\usepackage{algorithm} 
\usepackage{algpseudocode} 
\usepackage{arydshln} 

\graphicspath{{images/}}
\usepackage{makecell, booktabs}

\renewcommand{\vec}[1]{\mathbf{#1}}

\renewcommand{\vec}[1]{\mathbf{#1}}

\providecommand{\keywords}[1]
{
  \small	
  \textbf{\textit{Keywords---}} #1
}

\newcommand{\lbl}{l}

\makeatletter
\newcommand{\printfnsymbol}[1]{%
  \textsuperscript{\@fnsymbol{#1}}%
}
\makeatother

\title{
Mind The Gap: Modelling Difference Between Censored and Uncensored Electric Vehicle Charging Demand  
}
\author{Frederik Boe H{\"u}ttel, Filipe Rodrigues and
  Francisco C\^amara Pereira}

\begin{document}

\maketitle

\begin{abstract} 
Electric vehicle charging demand models, with charging records as input, will inherently be biased toward the supply of available chargers. These models often fail to account for demand lost from occupied charging stations and competitors.
The lost demand suggests that the actual demand is likely higher than the charging records reflect, i.e., the true demand is latent (unobserved), and the observations are censored.
As a result, machine learning models that rely on these observed records for forecasting charging demand may be limited in their application in future infrastructure expansion and supply management, as they do not estimate the true demand for charging. 
We propose using censorship-aware models to model charging demand to address this limitation. These models incorporate censorship in their loss functions and learn the true latent demand distribution from observed charging records. 
We study how occupied charging stations and competing services censor demand using GPS trajectories from cars in Copenhagen, Denmark. We find that censorship occurs up to $61\%$ of the time in some areas of the city.
We use the observed charging demand from our study to estimate the true demand and find that censorship-aware models provide better prediction and uncertainty estimation of actual demand than censorship-unaware models. We suggest that future charging models based on charging records should account for censoring to expand the application areas of machine learning models in supply management and infrastructure expansion.
\end{abstract}

\keywords{Electric mobility, Electric Vehicle charging demand, Latent mobility demand,  Bayesian modelling.}

\section{Introduction}
Electric vehicles (EV) and charging stations have become increasingly prominent in energy and transportation infrastructure planning \citep{jakobsen2020saadan, bauer2021charging}.
Adequate infrastructure is a crucial driving factor for large-scale EV adoption \citep{zonggen2018data, murugan2022elucidating}. However, the planning and expansion of charging infrastructure is a complex task that requires consideration of various factors such as charging station placement, demand forecasts, and integration with existing power systems and road infrastructure \citep{sanchari2021machinelearning}.
Machine learning has emerged as a promising tool to address some of the challenges associated with EV charging infrastructure planning, particularly charging demand prediction, which is well-suited for machine learning approaches \citep{buzna2019electric}.
Forecasts of the charging demand can provide energy providers and charging station operators with insights into the energy requirements of the current stations, aiding in operations and maintenance \citep{zhiyan2022electric}.

\begin{figure}[tb]
\includegraphics[width=0.45\textwidth]{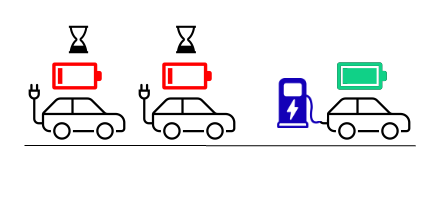}
\hfill
\includegraphics[width=0.45\textwidth]{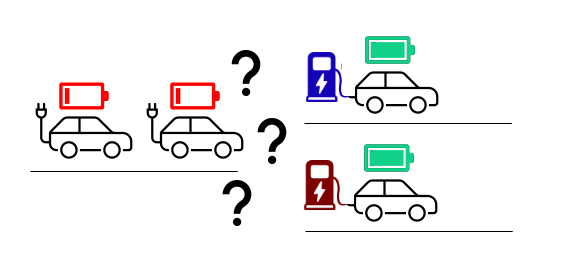}
\caption{The figure shows the two instances of censoring of electric vehicle (EV) charging demand Left: The demand is censored due to lost opportunities, where EV drivers can not charge due to occupied charging stations. Right: The demand is censored due to competing services, where EV drivers can choose between competing services (Blue and Red chargers).}
\label{fig:censoring}
\end{figure}

However, one distinguishing feature of charging stations is their shared nature, which affects how the charging demand should be modelled. Shared mobility services, like charging stations, often experience censorship, where demand is lost due to limited supply or competing services \citep{huttel2022modeling}.
Consequently, observations from charging stations may not accurately reflect the behaviour of service users, resulting in biased data and the subsequent models \citep{gammeli2020estimating}.
Specifically, for a charging station with two plugs, the station can never record a demand above two cars charging, even though more cars may need to recharge. The capacity of a charging station acts as an upper limit for the demand each station can observe, resulting in data that only contains a fraction of the true unobserved (latent) demand distribution. \autoref{fig:censoring} illustrates these scenarios.

Machine learning models trained with maximum likelihood, such as Mean Squared Error or Mean Absolute Error, on the observed data of EV charging demand do not account for the unobserved demand, which limits their ability to predict future demand for additional charging stations. 
This can lead to underestimating the true demand and, in turn, restrict operators' ability for strategic expansion and operational decisions. 
In contrast, censorship-aware models can provide estimates of actual charging demand by incorporating censorship into their loss functions and learning the true latent demand distribution from observed charging records \citep{huttel2022modeling}. By doing so, these models are better equipped to model frustrated demand from occupied charging stations and lost demand to competing service providers, resulting in more accurate demand forecasts for infrastructure planning.

Therefore, we propose to extend current models to be censorship-aware to estimate the true latent charging demand.
By incorporating censorship-aware techniques into the development of EV charging demand models, we can improve the accuracy and reliability of these models for infrastructure planning and policy making.
Our study concentrates on the two scenarios where censorship of demand may occur: firstly, when demand is lost due to occupied charging stations (frustrated demand) \citep{larsen2016demand}, and secondly, when competing services diminish the demand for a given provider \citep{chaojie2021data}.
Through a counterfactual analysis of demand in Copenhagen, Denmark, we provide insights into the challenges and limitations of traditional machine learning models when censorship is ignored and demonstrate how censorship-aware models can mitigate these issues.
In summary, the main contributions of the work are:
\begin{itemize}
    \item Previous research on electric vehicle demand modelling has mostly relied on observational records that do not account for frustrated demand and demand lost to competing services. In this study, we extend the models to be censorship-aware, which has the advantage that it can be applied to observed charging records to estimate the true latent demand.
    \item To analyse and validate the performance of our censorship-aware model, we conduct a counterfactual study on the charging demand in Copenhagen, Denmark, by modelling the censorship of various queue models and competing services. 
    \item We demonstrate the advantages of incorporating censorship into the EV charging demand models, particularly in scenarios where competing services and queue models are present. Specifically, our findings highlight the potential benefits of using these models in expanding and operating charging infrastructure.
\end{itemize}

We organise the rest of the work as follows. Section \ref{sec:litreview} reviews related work on modelling EV charging demand and censored modelling in the transport and machine learning domain. After the review, section \ref{sec:method} introduces two censorship-aware models with different distributional assumptions of the charging demand. Firstly, the Tobit model assumes a Gaussian demand distribution, and secondly, a quantile regression approach offers a semi-non-parametric distribution fit of the demand. 
It also covers how to model the spatial and temporal correlations between stations with graph neural networks. 
Section \ref{sec:data} introduces the data we base our models on, and section \ref{sec:experiments} describes our experimental setup, where we compare censorship-aware models with unaware models. In section \ref{sec:discussion}, we discuss some of the advantages and limitations of our modelling approach. Lastly, section \ref{sec:conclusion} concludes our main findings and outlines future research directions.

\section{Related Work}
\label{sec:litreview}
Over the last few years, there has been a growing interest in developing accurate models for predicting EV charging demand. Researchers have studied this problem from various perspectives, including simulation-based modelling \citep{jin2023electric}, queuing theory \citep{rich2022costbene}, and statistical models \citep{amara2022benchmark}. 
In this literature review, we focus on machine learning and deep-learning models for modelling the EV charging demand and highlight how observational data is used to model the demand. 
Additionally, we delve into censored modelling techniques and how they are applied to problems in transportation.

\subsection{Electric Vehicle Charging Demand.}
Generally, there is a large variation in the models used to model and forecast the EV charging demand. 
Traditional statistical models, such as the Autoregressive Integrated Moving Average (ARIMA) and its variations, have been widely employed to forecast the EV charging demand, as they can model the temporal correlations of the demand \citep{louie2017timeseries, amini2016arima, kim2021forecasting}.
These models offer interpretable parameters but may lack flexibility and non-linear fitting capabilities. As a result, various machine learning models, such as random forest \citep{almaghrebi2020data, lu2018random, Ullah2021electric}, support vector machines \citep{xydas2013forecasting, majidpour2016forecastin, sun2016chargingload, almaghrebi2020data}, and gradient boosting \citep{almaghrebi2020data, buzna2019electric}, have been used to forecast EV charging demand. These models often incorporate temporal and external features, such as weather, to enhance their predictive performance.

In recent years, recurrent neural networks (RNN) have become popular in demand forecasting due to their ability to handle large data sets and learn a signal's temporal features \citep{zhiyan2022electric}. One widely used RNN variant for forecasting EV demand is the long short-term memory (LSTM) \citep{juncheng2019electric, kim2021forecasting, kriekinge2021dayahead, taiyu2022multistep, boulakhbar2022deeplearning}. The LSTM is known for its non-linear fitting capabilities. 
However, its application is limited to the temporal aspect of the charging demand, as it fails to capture the complex spatial correlations between individual charging stations. \citet{zhiyan2022electric} propose to mitigate this limitation by clustered nearby charging stations and forecast demand at each cluster with a sequence-to-sequence LSTM architecture.
Another approach is to apply temporal graph neural networks, which combine recurrent neural networks and graph convolutions to incorporate spatial and temporal features into the forecast \citep{huttel2021deep}. A particular advantage of the graph neural network is that it can capture demand patterns at different spatial resolutions instead of at an aggregated level \citep{chaojie2021data}, and it learns the complex correlation between spatial locations \citep{tygesen2022unboxing}. However, the application of graph neural networks in EV charging demand modelling is still sparse.

To train and fit charging demand models, researchers often rely on \emph{observed} records from charging stations \citep{flammini2019stats, kim2021forecasting, kriekinge2021dayahead, buzna2021anensemble, taiyu2022multistep, boulakhbar2022deeplearning}. For example, \citet{amara2022benchmark} used public charging records from multiple cities to estimate demand, while \citet{shiyam2022electri} leveraged charging records to validate their PageRank-based demand predictions.
While charging records are an efficient way to estimate energy and power demand for charging stations, these models and data have limitations.
Specifically, observed charging records may not reflect the true demand due to limited supply and competition from alternative charging services. As a result, these models may not be suitable for infrastructure expansion planning as they cannot accurately extrapolate beyond the observed supply and estimate the latent demand.
The issue of censoring is often neglected in the research on EV demand forecasting. Therefore, in contrast to previous research,  we propose to extend the model to be censorship-aware to make more accurate forecasts of the true latent demand for charging.

\subsection{Censored Modelling of Mobility demand}

The modelling of censored observations is a well-established field, which includes models such as the Tobit model \citep{tobin1958estimation} and the censored quantile regression \citep{power1986censored, yu2007bayesian}.
As with the EV charging demand models, these traditional models are typically linear but recently have been extended to non-linear variations such as censored SVM \citep{shim2009support}, Random Forests \citep{li2020censored}, and lately neural networks \citep{Jia2022deepquantret}.

In mobility research, Gaussian processes (GP) have been the go-to model for censored distribution modelling \citep{gammeli2020estimating, gammeli2022generalized}. The GP assumes that the latent demand follows a Gaussian distribution and uses a Tobit likelihood function to provide a probabilistic fit.  \citet{gammeli2020estimating} used the GP to model the latent demand for bike-sharing systems and showed that censorship-aware models better fit the demand for shared bikes. \citet{xi2023censored} uses the censored GP to model the latent demand for bike-sharing systems and use the predictions for efficient resource allocation in bike-sharing systems. 
However, the GP faces scalability limitations when the data set size becomes too large \citep{Liu2020WhenGP} and the Gaussian assumption might be too simplistic for dynamic demand profiles \citep{huber2020probabilistic}.

As an alternative, neural networks can model the censored demand distribution, particularly with censored quantile regression models \citep{huttel2022modeling}.
The censored quantile regression is not limited to the estimation of a few moments, such as the mean and standard deviation, 
allowing it to fit non-symmetrical distributions \citep{peled2019preserving}.
For instance, \citet{huttel2022modeling} proposed using neural networks and censored quantile regression models to model shared mobility demand, including shared electric vehicle use. They proposed a censored quantile regression model, which provided a better uncertainty estimation than a Tobit model.
Moreover, censored quantile regression neural networks have been applied to biomedical data analysis \citep{Jia2022deepquantret} and in \citet{Pearce2022Censored}, they studied the effect of censorship in survival analysis. 

Although censorship is a common issue in transportation research, there is currently a lack of machine-learning techniques for modelling and analysing censored EV charging demand in the literature.
To address this gap and build upon previous works, we propose to model the charging demand with censorship-aware models, allowing a more accurate estimation of the true latent demand for charging. Specifically, we extend censorship-aware neural networks to deep architectures, including graph neural networks, to better capture the complex relationships and dependencies among charging stations.

\section{Methodology}
\label{sec:method}
To extend models to be censorship-aware in both the temporal and spatial dimensions, we derive the loss functions for the Tobit and censored quantile regression models.
Since EV charging demand is influenced by various spatial factors, such as the location and availability of nearby charging stations and temporal factors, such as daily and seasonal patterns, we extend the loss functions to temporal graph neural networks. 
We start by introducing the notations for censored modelling.

\subsection{Censored modelling}
A data set $\mathcal{D}=\{(x_i,y_i)\}_{i=1}^n$ is said to be \emph{censored} if the target values ($y_i$) are clipped observations of a corresponding latent (true) value $y^*_i$.  For each observation, there exists a threshold value $\tau_i$, which makes observations either left-censored or right-censored (or both), such that:

\begin{equation}
    y_i = 
    \begin{cases}
    y^* \,, & y^*_i > \tau_i \\
    \tau_i \,, & y^*_i \leq \tau_i
    \end{cases}
    \text{ in left-censorship,}
\end{equation}

\begin{equation}
    y_i = 
    \begin{cases}
    y^* \,, & y^*_i < \tau_i \\
    \tau_i \,, & y^*_i \geq \tau_i
    \end{cases}
    \text{ in right-censorship.}
\end{equation}

The threshold value $\tau_i$ in censored modelling can be observed, unobserved, fixed, or stochastic \citep{Pearce2022Censored, huttel2022modeling}. In right censoring, $\tau_i$ acts as an upper limit for the observations, while in left censoring, it acts as a lower limit. The primary objective of censored modelling is to approximate the latent value $y^*$ by using the data set $\mathcal{D}=\{(x_i,y_i, \tau_i)\}_{i=1}^n$ and a model with parameters $\theta$. The model is denoted as $f_\theta$, where $\theta$ is a set of parameters learned through stochastic gradient descent on a loss function. 

In the context of EV charging demand modelling, the variable $y^*_i$ denotes the true demand for charging, whereas $y_i$ corresponds to the observed demand, and $\tau_i$ is the demand that the charging infrastructure can serve. If the demand exceeds this threshold $\tau_i$, it will be truncated at the value of $\tau_i$, resulting in right-censored EV charging demand. 
It is important to note that one can switch between right and left censoring by negating $y_i$ and $\tau_i$. In the following section, we focus on the right-censoring scenario.

\subsection{The Tobit model}
The \emph{Tobit} model is commonly used in censored modelling \citep{tobin1958estimation}. It modifies $\tau_i$ to an indicator variable $\lbl_i$ set to 1 if $y_i$ is censored and 0 otherwise. The model assumes that the latent value, $y^*$, follows a Gaussian distribution given by,

\begin{equation}
y^*_i=f_\theta(x_i)+\varepsilon_i, \quad \varepsilon_i \sim \mathcal{N}(0,\sigma^2)\,.
\end{equation}

The likelihood function for the Tobit model can be derived as follows:
\begin{itemize}
    \item If $\lbl_i=0$, then the observation is not censored, and the parameter can be estimated with normal maximum likelihood. The likelihood function is given by
    \begin{equation}
        \mathcal{L}(\theta,x_i) = \varphi(y_i|f_\theta(x_i))
    \end{equation}
    where $\varphi$ is the Probability Density Function of a Gaussian distribution.
    \item if $\lbl_i=1$, then $y_i$ is censored, and the likelihood function is given by 
    \begin{equation}
        \mathcal{L}(\theta,x_i) = 1-\Phi(y_i|f_\theta(x_i))\, ,
    \end{equation}
    where $\Phi$ is the Cumulative Density Function of a normal distribution \citep{gammeli2020estimating}.
\end{itemize}

Assuming that all observations are independent and we have observed $\lbl_i$, combining the two censorship cases provides the following likelihood for a censored data set $\mathcal{D}$,

\begin{equation}
    \mathcal{L}\left(\theta, \mathcal{D}\right) = \prod_{i\in \mathcal{D}} \left(
    \underbrace{\left(\varphi(y_i|f_\theta(x_i)
    \right)^{(1 - \lbl_i)}}_{\text{Uncensored part}} 
    \underbrace{\left(1-\Phi(y_i|f_\theta(x_i))\, 
    \right)^{\lbl_i}}_{\text{Censored part}}
    \right)
    \,.
\end{equation}

The likelihood function is split into two parts: the censored case and the uncensored case. In many cases, it is common to fit models with the log-likelihood instead of the likelihood itself,

\begin{equation}
\log \mathcal{L}\left(\theta,\mathcal{D}\right)=\sum_{i \in \mathcal{D}} \Big((1-\lbl_i)\log \left(\varphi\left(y_i|f_\theta(x_i) \right) \right)+\lbl_i\log\left(1-\Phi\left(y_i|f_\theta(x_i) \right)\right)\Big)\, .
\end{equation}

In the context of neural networks and deep learning, the models are extended to have two outputs. One output represents the mean $\mu_\theta(x_i)$, while the other represents the standard deviation $\sigma_\theta(x_i)$. The output for the standard deviation is passed through a SoftPlus function to ensure it stays positive and differentiable. 
This makes the modelling assumption that $y_i^* \sim \mathcal{N}(\mu_\theta(x_i), \sigma_\theta(x_i))$. The models' outputs are used as the parameters for $\varphi$ and $\Phi$

\subsection{Censored quantile regression}
An alternative to the Tobit model for estimating censored data is censored quantile regression \citep{power1986censored}. 
Rather than estimating the mean and standard deviation of the latent distribution, quantile regression estimates a set of quantiles $\hat{Q}_\theta(x_i)=\{\hat{q}_{\theta,k}(x_i)\}_{k=1}^K$, where $\hat{q}_{\theta,k}(x_i)$ is the estimated $k$-th quantile of $y^*_i$. Multiple quantiles can be combined to model the target distribution of $y_i^*$.
It is important to note that $k$ must be constrained to the $\left[0,1\right]$ range.

Quantile regression models estimate the conditional quantile of the target distribution $y$, and for a single quantile $q_{\theta,k}$, the model can be expressed as:

\begin{equation}
y^*_i=f_\theta(x_i) + \varepsilon_{q, i}
\end{equation}

where $\varepsilon_{q,i}$ is the $q$-th conditional quantile of $\varepsilon_{i}$ \citep{yu2007bayesian}. 

For an uncensored quantile regression model, fitting is done by minimising the tilted loss function,
\begin{equation}
    \label{eq:tl}
    \rho_q(e) = \max\{q e\,, (q - 1)e \}
    \,.
\end{equation}
with a targeted quantile $q$ and the residual $e$.
The tilted loss is essentially a modified version of the $\ell_1$ loss, where a tilting parameter $q$ is introduced.
When estimating a given quantile (e.g., $q =0.05$ quantile), the tilted loss $q$ penalises overestimation more than underestimation by scaling the absolute value of the residual with its probability $q$ \citep{rodr2020beyond}.
The parameters of a quantile regression model $\theta$ are estimated by minimising the tilted loss over the entire data set. The loss function becomes:
\begin{equation}
\mathcal{L}(\theta, \mathcal{D})=\sum_{i \in \mathcal{D}} \rho_q \left(y_i-f_\theta(x_i)\right) .
\end{equation}

For a censored quantile model with left censorship, we observe $y_i$ instead of $y_i^*$, such that $y_i=\max\left\{\tau_i, y^*_i\right\}$. 
If we want $f_\theta(x)$ to estimate the quantile of $y^*$, the loss function becomes \citep{yu2007bayesian}:

\begin{equation}
    \mathcal{L}\left(\theta, \mathcal{D}\right)
    = \exp\Bigg\{ -\sum_{i\in \mathcal{D}} \rho_q ( y_i - \max \{\underbrace{\tau_i}_{\text{Censored part}}, \underbrace{f_\theta(x_i)}_{\text{Uncensored part}}\} ) \Bigg\}
    \,, 
\end{equation}

It is important to note that since the likelihood functions above are based on left censoring, the target values should be negated to convert the left censoring scheme to a right censoring one, as follows:

\begin{equation}
    \mathcal{L}\left(\theta, \mathcal{D}\right)
    = \exp\Bigg\{ -\sum_{i\in \mathcal{D}} \rho_q (-y_i - \max \{-\tau_i, -f_\theta(x_i)\} ) \Bigg\}
    \,, 
\end{equation}

To estimate the entire set of quantiles $Q_\theta$, one can extend a neural network or deep learning model to have $K$ outputs corresponding to the number of desired quantiles \citep{huttel2022modeling}. The likelihood function for this case is given by:

\begin{equation}
    \mathcal{L}\left(\theta, \mathcal{D}\right)
    = \sum_{q\in Q}
    \exp\Bigg\{ -\sum_{i\in \mathcal{D}} \rho_q \left(-y_i - \max \left\{-\tau_i, -f_\theta(x_i)\right\} \right) \Bigg\}
    \,, 
\end{equation}
where $Q$ is the set of quantiles to estimate.

\subsection{Censored Temporal Graph Convolution Neural Networks}
Temporal Graph Convolution Neural Networks (T-GCNs) are deep learning models that can effectively model non-linear relationships in both spatial and temporal dimensions of time series data. By combining graph convolution and recurrent layers, T-GCNs are highly effective in extracting spatial and temporal features from data \citep{zhao2020tgcn}. 
 
The spatial structure in T-GCN is modelled with a graph $G=(N, E)$ with $m$ nodes. The set of nodes is represented by $V=\{v_1, v_2, \dots, v_m\}$, and the edges between the nodes are represented by the set $E$. 
In a graph neural network, the dimensions of $x_i$ are equivalent to the number of nodes in the graph ($m$). Hence, $x_i$ can be expressed as a vector of values $\mathbf{x}_i \in \mathcal{R}^{m \times 1}$. This can include external features such as weather or time of day.
In temporal modelling, the target variable $\mathbf{y}_i$ and the input variable $\mathbf{x}_i$ are often derived from the same signal $\mathbf{x}$. In such cases, lagged values of the target are used to forecast future values, e.g., we have $\mathbf{x}_i = \mathbf{x}_{t-l:t}$ and $\mathbf{y}_i=\mathbf{x}_{t+1}$, where $l$ lagged values are used to forecast the target. 
In the case of censored modelling, the forecasting process can be expressed as follows
\begin{equation}
    \mathbf{x}^*_{t+1}=f_\theta(G, \mathbf{x}_{t-l:t}) + \bm{\varepsilon}\, ,
\end{equation}

Where $\mathbf{x}^*_{t+1}$ is the future true latent demand for each node in the graph, while $\bm{\varepsilon}$ depends on the choice of the loss function (as discussed in the previous section). The model $f_\theta(G, \mathbf{x}_{t-l:t})$ takes as input the graph structure $G$ and a window of past observations $\mathbf{x}_{t-l:t}$ and produces a forecast for the next time step. 

\paragraph{Spatial Modelling}
T-GCNs leverage the topology of the charging stations to define the edge weights of the graph $G$. 
A common method to weight an edge between nodes $i$ and $j$ is via a Gaussian Kernel weighting function, defined as \citep{shuman2013emerging},

\begin{equation}
    e_{ij}= \exp(-h(z_i,z_j))\,.
\end{equation}

Here, $z_i$ and $z_j$ denote the latitude and longitude of the nodes, respectively, and $h$ represents the Haversine Distance between the nodes measured in kilometres \citep{huttel2021deep}. 
This makes nodes close to each other have a large weight, i.e. a strong connection in the graph.
Using the set of edges $E$, we can construct the adjacency matrix $A \in R^{m\times m}$, representing the network's topology.

To capture features from the topological structure, a graph convolutional network (GCN) uses graph convolutions to model the relationship between nodes in a graph \citep{kipf2016semi}. In the case of T-GCNs, a two-layered GCN model is often employed. The model is formulated as follows
\begin{equation}
\label{eq:GCN}
 f(A,\vec{x}_t)=\sigma\left(\widehat{A}\, \operatorname{Relu}\left(\widehat{A} \vec{x} W_{0}\right) W_{1}\right)\,,
\end{equation}

Here, $\widehat{A}=\widetilde{D}^{-\frac{1}{2}} \widetilde{A} \widetilde{D}^{-\frac{1}{2}}$ is the adjacency matrix with a self-connection structure (i.e. it has non-zero entries in the diagonal), and $\widetilde{D}$ is the degree matrix with $\widetilde{D}_{ii}=\sum_{j}\widetilde{A}_{ij}$. The weight matrices in the first and second layers are denoted by $W_0$ and $W_1$, respectively. The activation function $\sigma(\cdot)$ is applied element-wise, and $\operatorname{Relu}()$ is the rectified linear unit activation function \citep{zhao2020tgcn}.

\paragraph{Temporal Modeling}
The GCN extends to a temporal signal by combining the GCN with long short-term memory layers (LSTM) \citep{LSTM, zhao2020tgcn}. The key equations of the T-GCN with an LSTM cell can be summarised as follows, where $f(A,\vec{x}_t)$ is the graph convolution from \autoref{eq:GCN}:

\begin{align}
i_{t} &=\sigma_{g}\left(W_{i} f(A,\vec{x}_{t})+U_{i} h_{t-1}+b_{i}\right)\,, \\
f_{t} &=\sigma_{g}\left(W_{f} f(A,\vec{x}_{t})+U_{f} h_{t-1}+b_{f}\right)\,, \\
o_{t} &=\sigma_{g}\left(W_{o} f(A,\vec{x}_{t})+U_{o} h_{t-1}+b_{o}\right)\,, \\
\tilde{c}_{t} &=\sigma_{c}\left(W_{c} f(A,\vec{x}_{t})+U_{c} h_{t-1}+b_{c}\right)\, , \\
c_{t} &=f_{t} \odot c_{t-1}+i_{t} \odot \tilde{c}_{t}\,, \\
h_{t} &=o_{t} \odot \sigma_{h}\left(c_{t}\right)\,.
\end{align}

The trainable weights are contained in the matrices $W_{i-c}$ and $U_{i-c}$ and the vectors $b_{i-c}$. The initial values of the cell and hidden states are set to $c_0=0$ and $h_0=0$, respectively. The Hadamard product is denoted by $\odot$, the forget gate activation is represented by $f_t$, the input/update gate activation is represented by $i_t$, and the output gate hidden activation is represented by $o_t$. The Sigmoid function is denoted by $\sigma_g$, and the hyperbolic tangent activation function is represented by $\sigma_c$ and $\sigma_h$ \citep{LSTM}. 
Therefore, the set of parameters $\theta$ comprises the matrices of the weights matrices from \autoref{eq:GCN}, and all the weights in the LSTM ($W_{i-c}$ and $U_{i-c}$ and the vectors $b_{i-c}$)

In summary, the T-GCN models capture the complex topological structure using the GCN and the temporal structure of the data using the LSTM layers \citep{zhao2020tgcn}. A single forward pass produces a forecast for each node in the graph.

\paragraph{Censored Graphs}
We extend the model's architecture to integrate the censored likelihood functions to make the T-GCN censorship-aware. Specifically, we extend the output to provide either the mean and standard deviation of a Gaussian distribution  ($\mathcal{N}(\mu_{\theta}(x_i)), \sigma_{\theta}(x_i)$)  or the set of quantiles ($\hat{Q}_{\theta}(x_i)$) for each node in the graph. For simplicity in notation, the predicted values for time $t+1$ are denoted as $y_i$.
Each graph node's threshold value, $\tau_{i,v}$, depends on its corresponding supply and can vary across the nodes. 
The loss function is computed individually for each node, $v$, and then summed.
The loss function for the Tobit T-GCN becomes

\begin{equation}
\label{eq:tobit_fit}
\mathcal{L}\left(\theta,\mathcal{D}\right)=
\sum_{i \in \mathcal{D}} \sum_{v\in V}\Big((1-\lbl_{i,v})\log \left(\varphi\left(y_{i,v}|f_\theta(x_{i,v}\right) \right)+\lbl_{i,v}\log\left(1-\Phi\left(y_{i,v}|f_\theta(x_{i,v}) \right)\right)\Big)\, ,
\end{equation}

And the loss function for the censored quantile regression T-GCN becomes 

\begin{equation}
    \mathcal{L}\left(\theta, \mathcal{D}\right)
    = \sum_{q\in Q} 
    \exp\Bigg\{ -\sum_{i\in \mathcal{D}}\sum_{v\in V}  \rho_q \left( y_{i,v} - \max \left\{ \tau_{i,v}, f_\theta(x_{i,v})\right\} \right) \Bigg\}
    \,. \label{eq:cqrgcn}
\end{equation}

It is worth noting that the likelihood functions for both censored and censored quantile regression models are not specific to a particular neural network architecture and can also be utilised with other types of neural networks.
One notable advantage of censored models is their ability to handle observed charging records, given that the information regarding which observations are censored is available. This property enables the approximation of the true charging demand, even when the observed demand is incomplete due to censorship.

\section{Counterfactual study}
\label{sec:data}
\begin{figure}[tb]
    \centering
    \includegraphics[width=\textwidth]{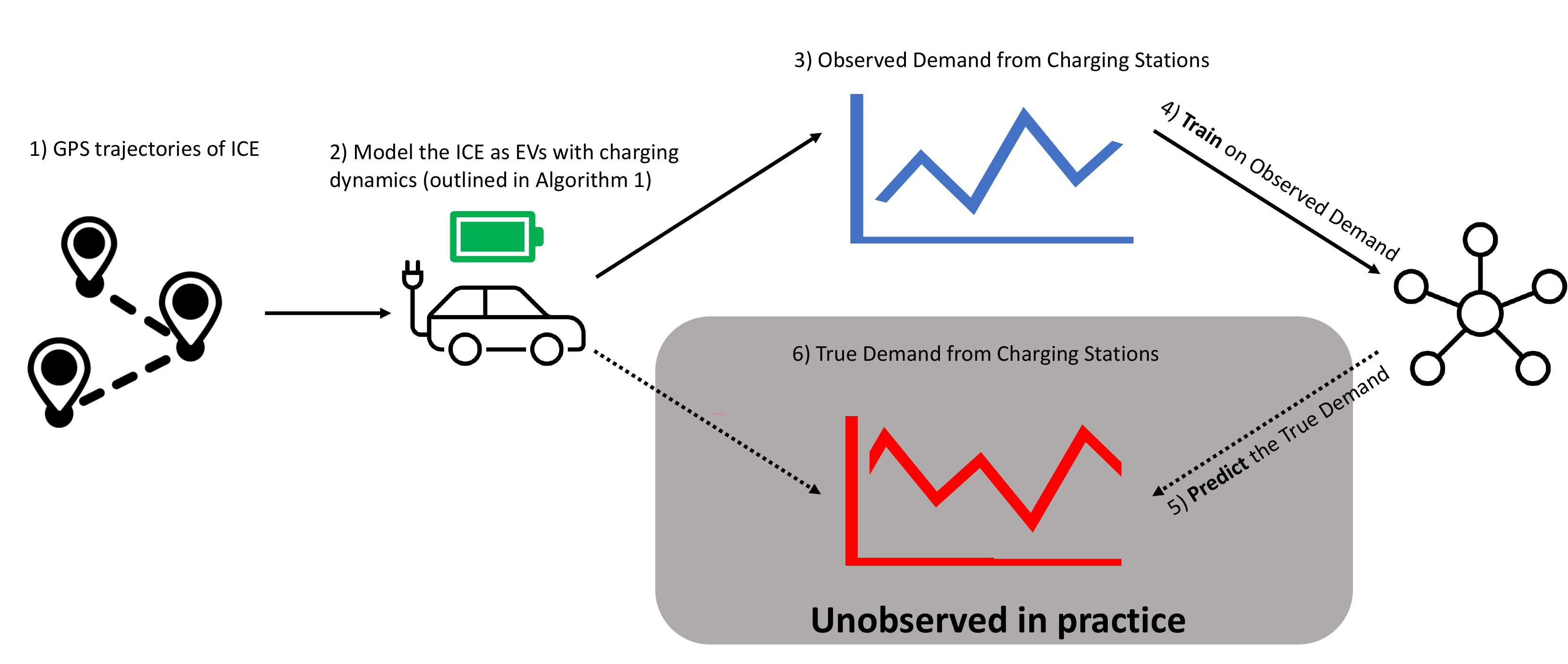}
    \caption{Illustration of the experimental procedure. 
     Firstly, the GPS trajectories of Internal Combustion Engine (ICE) vehicles are modelled as EVs using Algorithm~\autoref{alg:overview} (for detailed information, refer to \autoref{sec:counterfactualstudy}). Subsequently, the observed charging demand and the true demand are derived from the study. The observed demand is utilised for training a censorship-aware T-GCN model, which is then employed to predict the true demand.
     In practice, the true demand is not directly observable because of censoring.}
    \label{fig:overview}
\end{figure}

To test the hypothesis that observed EV charging demand is censored, we conduct a \emph{counterfactual} study using GPS trajectories of internal combustion engine cars (ICE) in the metropolitan area of Copenhagen, Denmark. The ICE trajectories estimate both observed and unobserved charging demand in the city, assuming that the ICE cars are replaced with EVs. We model the state-of-charge (SoC) of the EV fleet based on the GPS trajectories and charging dynamics from public charging stations, similar to the approach proposed by \citet{TU2016172}. This study provides us with the \emph{true} unobserved charging demand, allowing us to assess the censorship levels that occur throughout space and time as we model each charging station with a queuing model. 
We call it a counterfactual study because we ask, "What would have happened with charging demand if the ICE cars from the data set had been EVs instead?" 
The overall procedure is illustrated in \autoref{fig:overview}, while Algorithm~\ref{alg:overview} provides a summary of the study's steps, which are further explained in \autoref{sec:counterfactualstudy}.
The study aims to obtain charging records for both charged and uncharged vehicles, enabling us to assess the censorship in EV charging demand and examine its spatial and temporal characteristics.


\begin{figure}[b]
    \centering
    \includegraphics[width=0.95\textwidth]{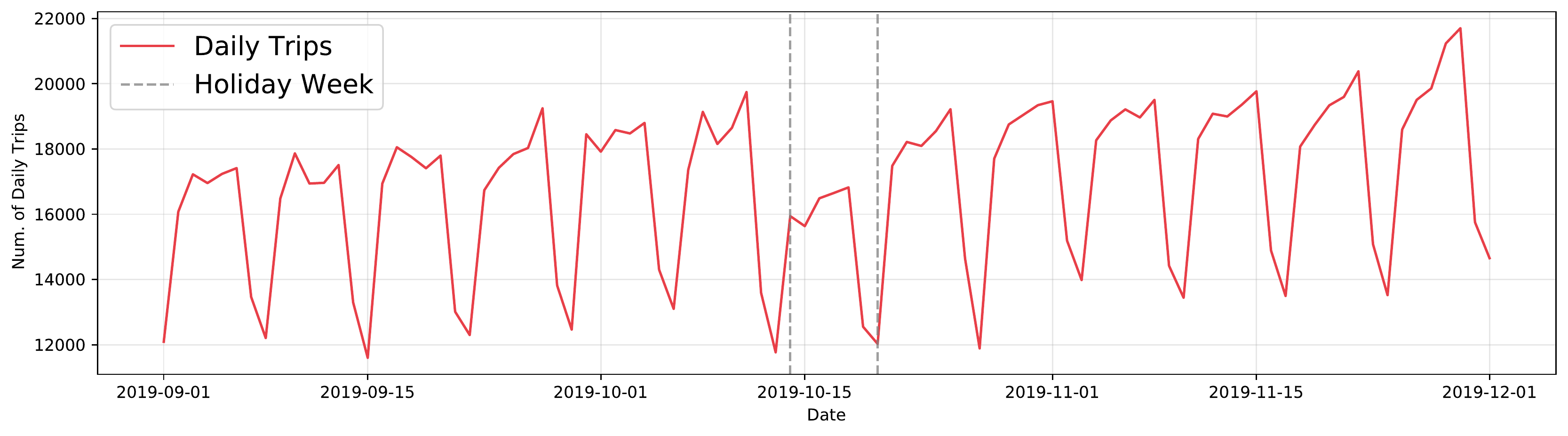}
    \caption{Overview of the total daily trips in the data set. From the 14th to the 20th of October, there is a drop in daily trips due to a holiday week in Denmark. Therefore, there are fewer trips due to a reduced traffic flow from commuting.}
    \label{fig:num_daily_trips}
\end{figure}

\subsection{Data}
The data consist of GPS trajectories from ICE cars in Denmark, where locations are saved every 20 seconds. 
Each GPS location contains an ID of the car and can be stitched together into a complete GPS trajectory, which forms a trip for the car.
Each trip contains a start coordinate, an end coordinate, and the distance driven by a car.
Cars' parking time can be inferred as the time between trips.
To ensure the users' privacy, the trajectories are randomised by adding a noise distance between 50 and 500 meters to each trip's endpoints. 
The GPS trajectories of the cars were collected over three months, from September to November 2019 and are uniformly distributed across the country (Denmark) and vehicle segments. 
The GPS trajectories do not suffer from the biases identified in other EV studies, namely, having early adopters bias the behaviour in the data sets.
In total, 32664 cars are observed in the capital region (Copenhagen), which accounts for $5.71\%$ of the cars in the city (A total of 571627 cars for the period \citep{statbank2022}). 
In the observations period, the  penetration rate of EVs in Copenhagen was $1\%$, and for the fall of 2022, the penetration rate has increased to $2.5\%$~\citep{statbank2022}. 
We visualise the number of daily trips for the entire period in \autoref{fig:num_daily_trips}. 
In total, we have 1.5 million trips with a median of 10 trips pr. Car.

\paragraph{Charging Stations in Copenhagen}
In this study, the charging infrastructure in Copenhagen was modelled based on the charging infrastructure from 2021. To obtain the locations of EV charging stations, we scraped the open-source map provided by \citet{uppladdning_2021}. The scraped chargers have varying charging power ranging from slow chargers with 3.7 kW to fast chargers with 150 kW. The extracted information provides us with the locations and charging power for each charging station in the city. This information was then used to simulate the locations where ICE cars would have charged had they been EVs.

\subsection{Queuing Models}
We use three different queuing methods to handle multiple cars wanting to charge. We theorise that the observed demand is censored, as opportunities are lost when charging stations are occupied. We model three different queues, which all have real-world applications. 
The three queues are as follows:

\begin{itemize}
    \item \textit{Gas station}: Cars queue up like an ordinary gas station and wait for their turn to charge. The EVs charge to 80\% SoC before driving away. This behaviour is primarily associated with fast chargers and is often called a smart queue \citep{Monta}.
    \item \textit{3-Hour}: In this queue, EVs are only allowed to charge for 3 hours before driving away, regardless of the SoC of the car. Cars that arrive at an occupied station will not be charged. This behaviour reflects policies (e.g. Denmark \citep{FDM}) where there is a (typically 3 hrs) time limit for parking and charging.
    \item \textit{First comes - First serve queue}: The last queue follows a first come, first serve approach, where the EV that arrives first charges for the entire duration until the next trip. Cars that arrive at an occupied station will not be charged. This corresponds to opportunistic behaviour.
\end{itemize}

The different queuing methods proposed in our study result in varying levels of censoring. For example, the gas station queue can accommodate all users as long as they are willing to wait. In contrast, the first come - first serve queue can only serve users who find an available charger. We maintain consistency in our queuing approach across all charging stations for one study.

\subsection{Outline of the Counterfactual Study}
To study the charging behaviour of ICE cars as if they were EVs, we randomly sample different numbers of cars from the GPS trajectories, corresponding to different EV penetration rates. The GPS trajectories originally have a penetration rate of $5\%$ for all cars. However, since the penetration rate of EVs was roughly $1\%$ in 2019 and $3\%$ in 2022, we vary the penetration rate from $1\%$ to $5\%$ and test the three different queuing models.
We model the cars' charging behaviour using the sampled cars and the queuing model, assuming they were EVs. We randomly sample the battery sizes according to the market distribution of different EV types as described in \autoref{sec:marksetshareev}. We also calculate their willingness to charge based on the state of charge consumption of a trip.
The charging station is probabilistically selected from each end trip's five closest charging stations. Each car follows the procedure in Algorithm \ref{alg:overview}.

\begin{algorithm}
\caption{Overview of the counterfactual study.}\label{alg:overview}
\begin{algorithmic}
\State Generate a fleet of electric vehicles.
\State Set queuing model.
\State Set SoC for the fleet of cars $SoC \sim \mathcal{N}(0.6,0.2), 0.20 \leq SoC \leq 1$.
\For{each GPS trajectory}
        \State Find the car id of the GPS trajectory.
        \State Compute energy consumption for the car.
        \State Update SoC-level for the car.
        \State Compute willingness to charge for the car.
        \State Compute were to charge based on the end location of the trip.
        \State Add car to queue.
        \State Compute energy demand based on the time spent charging.
        \State Save the energy demand from the charging event.
\EndFor
\end{algorithmic}
\end{algorithm}

\subsection{Demand Profiles from the Counterfactual Study} 
We examine the resulting observed and unobserved energy demand from each GPS trajectory.
We find that censorship occurs and varies across the different queues and penetration rates. The gas station queue experiences very little censorship due to the nature of charging, where observations are only censored if the EVs have a new trip while in the queue. However, the censoring increases as we move to the different queue types, with the first-come queue having the largest amount of censorship.
We report the fraction of hours where censorship occurs across the queue and the penetration rate in \autoref{tab:hours_censoring}.
For the current penetration levels of EVs, we find that between $33\%$ and $53\%$ of hours are censored in the entire area. As the penetration rate increases, censorship follows suit, which is a natural consequence, as the infrastructure would need to expand to handle more demand from chargers.

\begin{table}[tb]
    \centering
    \resizebox{0.95\textwidth}{!}{
    \begin{tabular}{l|rrrrr}
    \toprule
    & \multicolumn{5}{c}{Penetration rate}  \\
    \thead{Queue} & \thead{1\% (Historical)}& \thead{2\%} & \thead{3\% (Current)}  & \thead{4\%} & \thead{5\% (Data)} \\
    \midrule
    Gas station Queue &   $<$0.1\% & 2.2\% & 4.9\% & 9.8\% & 16.4\% \\
    3 Hour queue &  4.1\% & 17.6\% & 33.2\% & 46.8\% & 56.9\% \\
    First come - First serve Queue & 8.3\% & 30.1\% & 53.2\% & 66.2\% & 70.5\% \\
    \bottomrule
    \end{tabular}
    }
    \caption{Fraction of hours where the observed demand is censored across the different queuing models and penetration rate.}
    \label{tab:hours_censoring}
\end{table}

\begin{figure}[tb]
\includegraphics[width=0.495\textwidth]{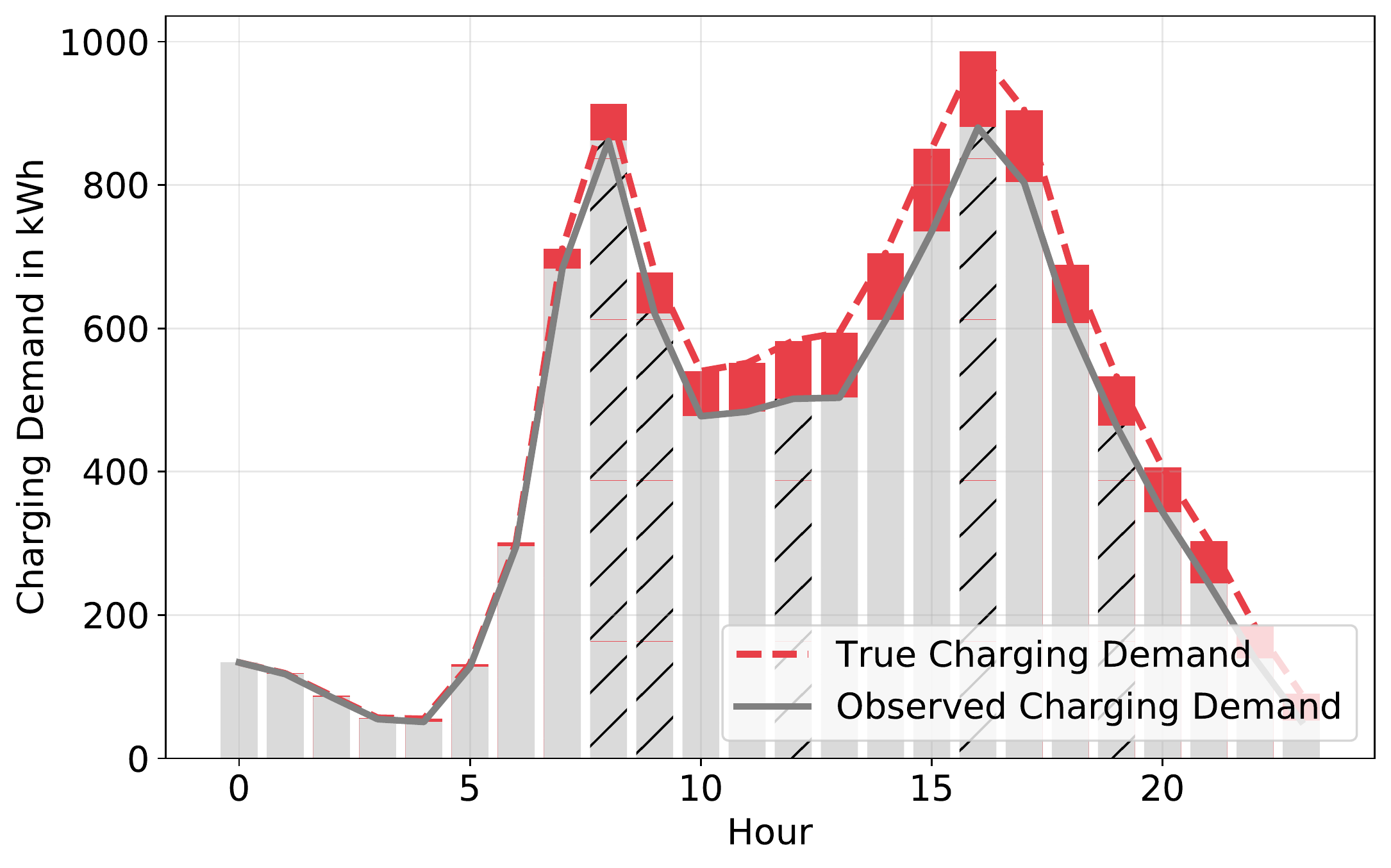}
\hfill
\includegraphics[width=0.485\textwidth]{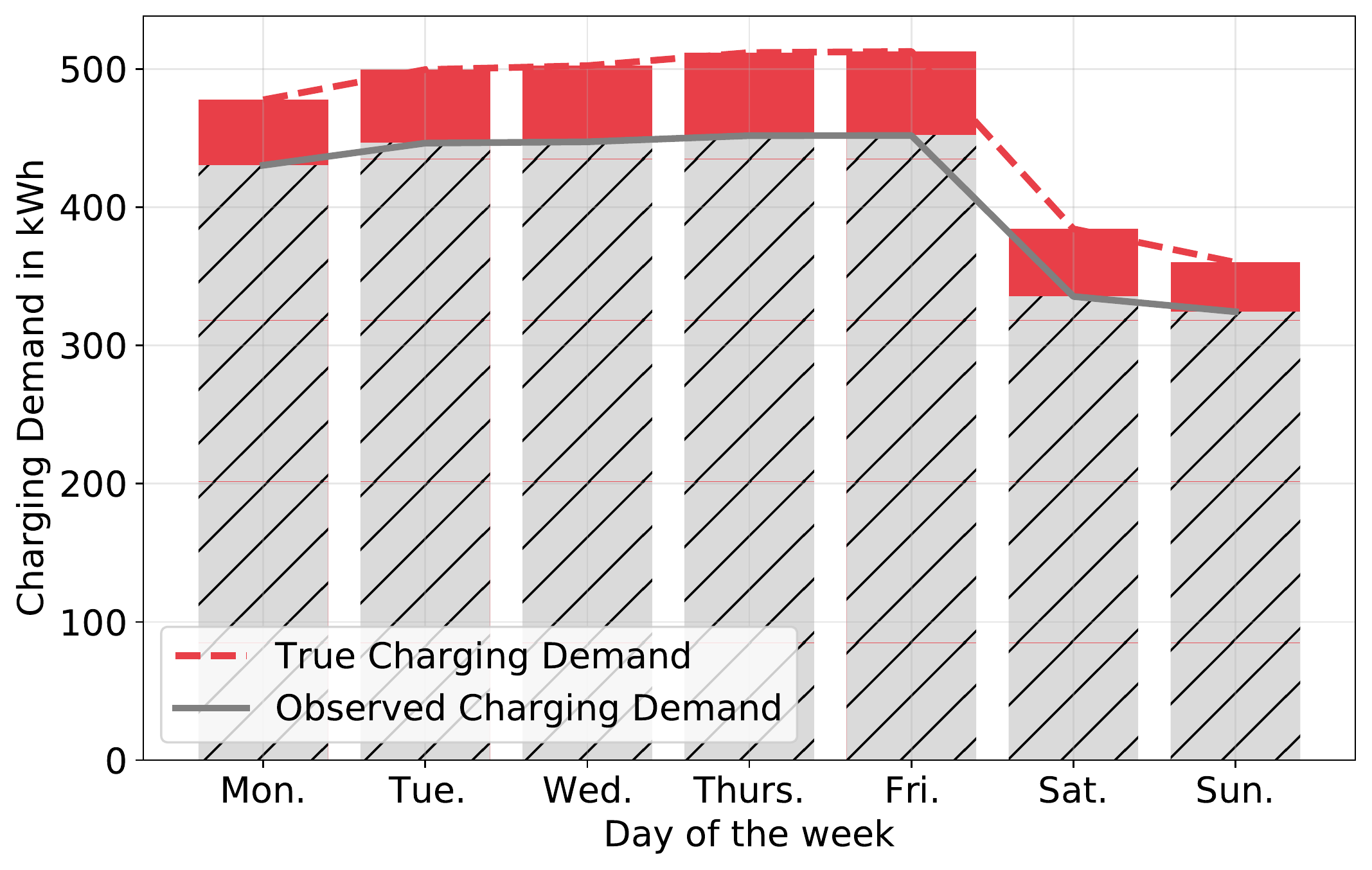}
\caption{Average charging demand, either grouped by the time of day (left) or day of the week (right). Rate: $5\%$ and queue: \emph{First come - First serve}.}
\label{fig:averages}
\end{figure}

The study shows that most censoring in EV charging demand occurs during peak demand hours in the temporal dimension, particularly during morning and afternoon periods, as shown in \autoref{fig:averages}. 
These peak demand hours align with the traffic patterns observed in the original GPS trajectories and are consistent with findings from other studies on charging demand in urban areas. Furthermore, the analysis revealed that charging demand varies across different days, with lower demand observed on weekends than on typical weekdays.
Notably, despite the fluctuations in the total charging demand, the amount of censored demand remains relatively stable, with no significant variations in the proportion of censored demand observed across different weekdays (\autoref{fig:averages}). These findings provide important insights into the temporal patterns of EV charging demand and highlight the need for censorship-aware models to accurately estimate the true latent demand distribution, especially during peak demand periods.

\begin{figure}[tb]
    \centering
    \includegraphics[width=0.65\textwidth]{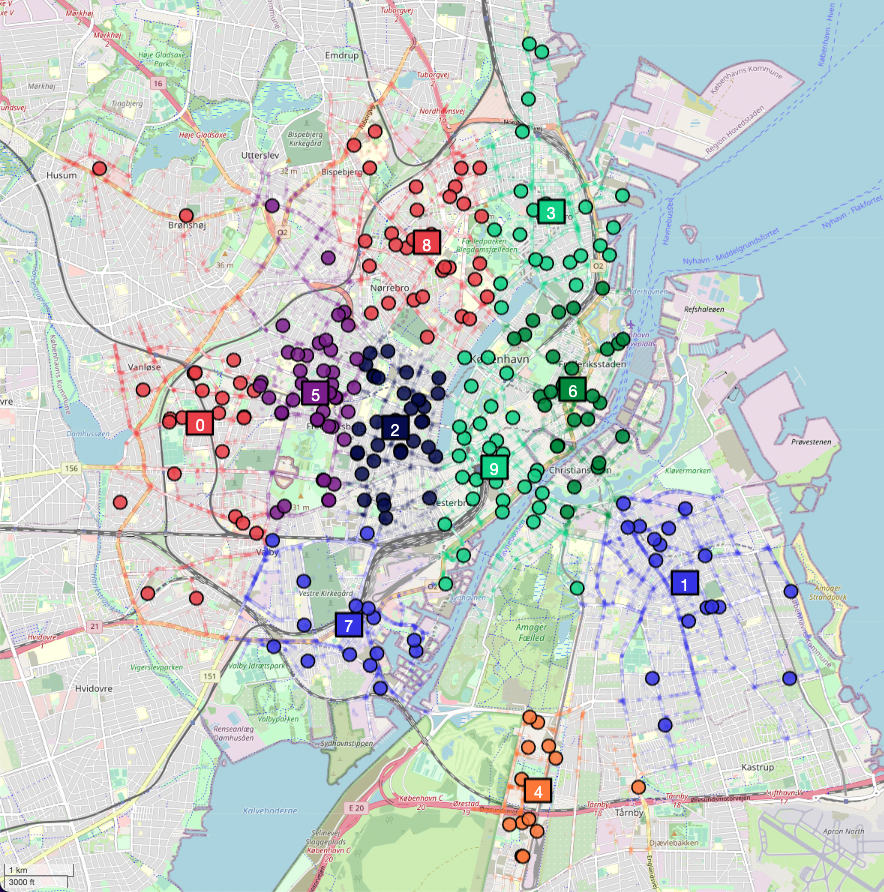}
    \caption{Overview of the spatial structure in the counterfactual study. Small circles indicate the end of the GPS trajectories. The circle with a dark line is the charging station, and squares indicate the centroids of the clusters. Colours indicate which cluster they belong to.}
    \label{fig:charging_stations_locations}
\end{figure}

\begin{table}[tb]
    \centering
    \begin{tabular}{llr}
    \toprule 
    \thead{Cluster} &\thead{Location} &\thead{$\%$ of censored hours}  \\  \midrule
    0 & Western part of city & $61.03\%$ \\
    1 & South Easter part &  $20.29\%$ \\
    2 & West of the city centre & $3.12\%$ \\
    3 & North Eastern part of the city & $16.31\%$ \\
    4 & Southern part of the city & $11.28\%$ \\
    5 & Western part of city & $24.33\%$ \\
    6 & North part of the city centre & $6.57\%$ \\
    7 & South Western part of the city & $20.16\%$ \\
    8 & North Western part of the city & $31.17\%$ \\
    9 & City center & $16.67\%$ \\
    \bottomrule
    \end{tabular}
\caption{Table presenting the proportion of hours with censored observations under a $5\%$ EV penetration rate and the First come - First serve Queue}
    \label{tab:censoringr05}
\end{table}

\begin{figure}[tb]
    \centering
    \includegraphics[width=\textwidth]{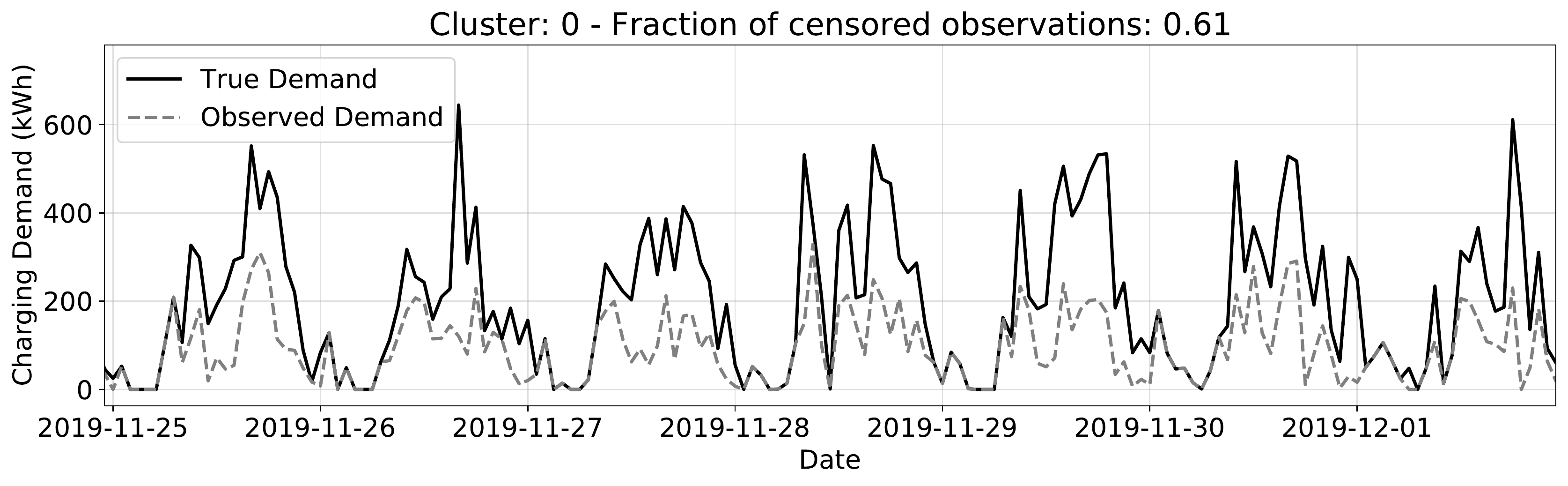}
\hfill
    \includegraphics[width=\textwidth]{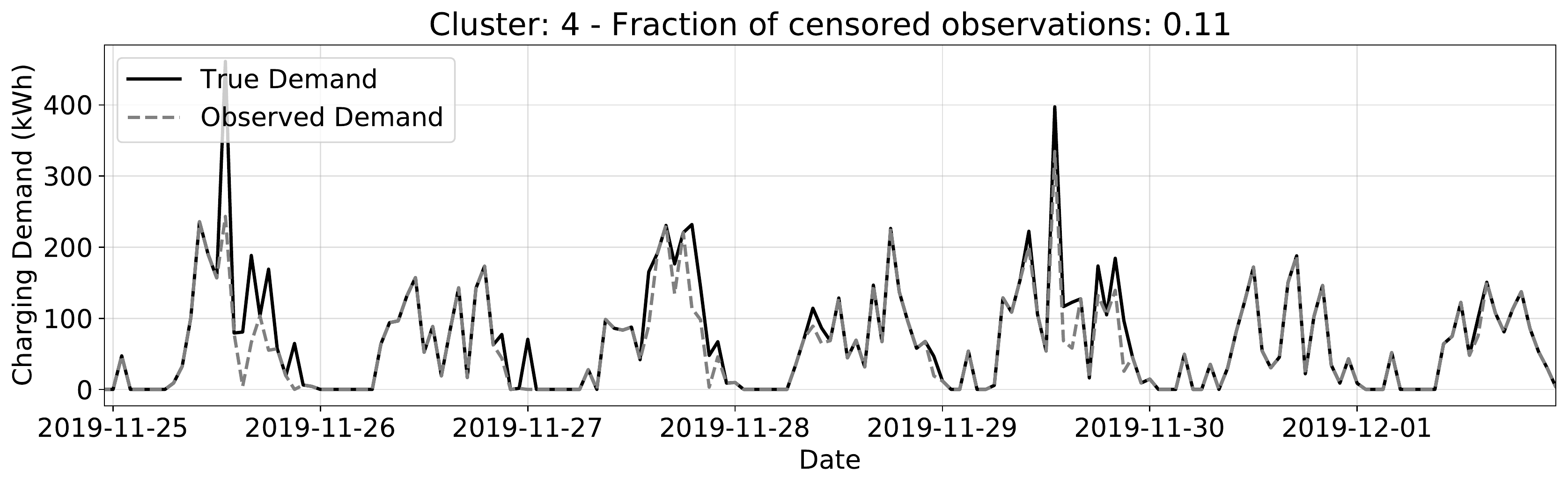}
    \caption{This figure displays the charging demand obtained from the Counterfactual study for cluster 0 and cluster 4 in the city. The observed demand is input into the models to estimate the true latent demand.}
    \label{fig:charging_stations_demand}
\end{figure}

We spatially aggregate  the demand data with k-means clustering, which groups charging stations into ten regions. This approach is necessary as demand at individual charging stations can be highly unpredictable and sporadic, making it difficult to model accurately. By clustering stations based on their proximity, we account for regional variations in demand patterns.
Each cluster centre serves as a node in the temporal graph neural network
To visually represent the distribution of charging stations used in our study, we present a map in \autoref{fig:charging_stations_locations}.
We find that across the city, the censorship levels vary. 
\autoref{tab:censoringr05} shows the censorship levels for the $5\%$-penetration of EVs and the \emph{First come - first serve} queue. In cluster 0, which covers the western part of Copenhagen, $61\%$ of the hourly charging demand is censored. This high level of censorship is attributed to only two regional charging stations covering a large spatial area without having sufficient supply to meet the demand. However, it is important to note that this area is primarily residential, and most of the demand could be served by home charging, which is not included in the study.
On the other hand, in cluster 4, located in the southern part of Copenhagen and an industrial area with low demand at night, lower censorship levels and minimal demand at night are observed (\autoref{fig:charging_stations_demand}). The study reports censorship levels across all the different queues, penetration rates, and clusters in \autoref{sec:censoring_across}.

To summarise, we conducted a counterfactual study to model the charging behaviour of electric vehicles (EVs) based on GPS trajectories from internal combustion engine (ICE) cars in the fall of 2019. 
This approach enabled us to estimate the true unobserved charging demand that would have existed had these cars been EVs. We find that the demand is censored and varies in space and time. 
The charging demand for all the cars, including those who recharged and those who did not, is obtained in the form of charging records similar to those obtained by charging station operators. The observed charging records are used as input to the Censored T-GCN models. The unobserved charging records are utilised to validate the model's accuracy in approximating the true charging demand.

\section{Experiments and Results}
\label{sec:experiments}
We conduct a series of experiments with the counterfactual study data. First, we investigated the effect of various censorship schemes on modelling the charging demand for the entire city while considering different penetration rates and queuing models. Next, we compared and evaluated the performance of censorship-aware and censorship-unaware models in a competitive setting where only a fraction of the demand is observable \footnote{Code is available at \href{https://github.com/fbohu/CensoredTGCN}{https://github.com/fbohu/CensoredTGCN}}.

We label the models based on the likelihood functions they employ. We include two censorship-unaware models: a \emph{Gaussian} model estimated using maximum likelihood and a \emph{quantile} regression (QR) model trained with the uncensored tilted loss function described in \autoref{eq:tl}. Additionally, we compare these models with censorship-aware models: a \emph{Tobit} model (\autoref{eq:tobit_fit}) and a \emph{censored QR} model (\autoref{eq:cqrgcn}). We adopt identical architectural choices for all models and initialise their parameters with the same random seed.
We conduct ten runs for each experiment and report the average results across these runs, along with their corresponding standard deviations.

\subsection{Architectures}
The most recent 168 hourly demand observations, representing a week's worth of data, are utilised as the input size for the temporal signal. To augment the historical demand data, the type of day and hour are incorporated as external input features encoded into cyclical features using sine and cosine. These cyclical features are combined with the historical demand to generate time series for each node in the graph. To prevent any one node from contributing excessively to the loss function during training, each time series is scaled between 0 and 1, as recommended by \cite{zhao2020tgcn}.
The model uses 16 and 8 channels for the graph convolutions and an LSTM with 32 hidden units. The Adam optimiser \citep{kingma2019adam} with a learning rate of 0.0003 and norm clipping of 1.0 is used. During training, each model is trained for 1000 epochs with a batch size of 256, and an early stopping criterion of 0.001 on the validation loss is applied.
To evaluate the models, the data set is divided into three sets: train, validation, and test, with a split of 80\%, 10\%, and 10\%, respectively. The censored quantile regression models estimate the 0.05, 0.5, and 0.95 quantiles. All models are implemented in Keras \citep{chollet2015keras}, and graph convolutions were performed using StellarGraph \citep{StellarGraph}

\subsection{Evaluation Metrics}
We assess the performance of our models by computing their tilted loss on the test set ($\mathcal{D}_{\text{test}}$). We estimate their quantiles using the fitted distributions for the Gaussian and Tobit models. Moreover, we use the following metrics widely used to evaluate a probabilistic forecast: Interval Coverage Percentage (ICP) and Mean Interval Length (MIL). To simplify notation, we consider a single node in the graph.

\begin{align}
    \text{ICP} &= \frac{1}{|\mathcal{D}_{\text{test}}|}\sum_{i\in \mathcal{D}_{\text{test}}}\begin{cases}
    1 & \text{ if } \hat{q}_{\theta,0.05}(x_i) \leq y^*_i \leq \hat{q}_{\theta,0.95}(x_i)   \\
    0 & \text{otherwise}
    \end{cases} \label{eq:icp} \\
    \text{MIL} &= \frac{1}{|\mathcal{D_{\text{test}}}|}\sum_{i\in \mathcal{D}_{\text{test}}}\left(|\hat{q}_{\theta,0.05}(x_i)-\hat{q}_{\theta,0.95}(x_i) |\right) \label{eq:mil}
\end{align}

The objective of a probabilistic model is to achieve a high Interval Coverage Percentage (ICP) (\autoref{eq:icp}), close to 0.9, indicating that $90\%$ of the observations fall within the prediction interval, while maintaining a low Mean Interval Length (MIL) (\autoref{eq:mil}), which indicates that the model's predictions intervals are tight around the true value.

\subsection{Total Demand Predictions}
To assess the effectiveness of our models, we conducted initial experiments using the censoring scheme outlined in \autoref{tab:hours_censoring}, which was applied to all charging stations and nodes. The penetration rate of electric vehicles (EVs) was varied, with corresponding increases in censorship levels and modifications to the queuing model. Our findings are presented in \autoref{tab:tilte_loss_test}, where we report the tilted loss on the true demand for the test set."

Our findings suggest that the quantile regression-based models (QR and Censored-QR) generally perform better than the censorship-unaware models. For the Gas station queue, we did not observe any substantial distinction between the censorship-aware and unaware models, primarily due to the low levels of censorship in this queue type. However, as we varied the queuing model and censorship levels, we found that the performance discrepancy between the censorship-aware and unaware models increased. Specifically, the Censored QR model is the best-performing model for the 3-hour and First come queue, with high censorship levels.

\begin{table}[tb]
    \centering
    \resizebox{1\textwidth}{!}{
    \begin{tabular}{ll|rrrrr}
    \toprule
    \thead{Queue} & \thead{Penetration rate} & \thead{1.0\%}& \thead{2.0\%} & \thead{3.0\%}  & \thead{4.0\%} & \thead{5.0\%} \\
    \midrule
    \multirow{4}{*}{Gas station} 
& Gaussian & $0.796 \pm 0.001$ & $0.800 \pm 0.001$ & $0.709 \pm 0.001$ & $0.783 \pm 0.001$ & $0.714 \pm 0.001$ \\
& Tobit & $0.801 \pm 0.009$ & $0.800 \pm 0.007$ & $0.711 \pm 0.004$ & $0.775 \pm 0.003$ & $0.713 \pm 0.011$ \\ 
& QR & $0.656 \pm 0.003$ & $\mathbf{0.719 \pm 0.002}$ & $\mathbf{0.668 \pm 0.003}$ & $0.731 \pm 0.006$ & $\mathbf{0.680 \pm 0.004}$ \\ 
& Censored QR & $\mathbf{0.655 \pm 0.006}$ & $0.720 \pm 0.008$ & $0.670 \pm 0.004$ & $\mathbf{0.731 \pm 0.004}$ & $0.681 \pm 0.005$ \\
    \midrule
    \multirow{4}{*}{3-hour} 
& Gaussian & $0.767 \pm 0.002$ & $0.750 \pm 0.002$ & $0.810 \pm 0.002$ & $0.781 \pm 0.002$ & $0.798 \pm 0.002$ \\ 
& Tobit & $0.764 \pm 0.006$ & $0.750 \pm 0.002$ & $0.799 \pm 0.007$ & $0.762 \pm 0.005$ & $0.765 \pm 0.006$ \\ 
& QR & $0.644 \pm 0.004$ & $0.678 \pm 0.005$ & $0.756 \pm 0.003$ & $0.740 \pm 0.003$ & $0.763 \pm 0.002$ \\ 
& Censored QR & $\mathbf{0.641 \pm 0.006}$ & $\mathbf{0.673 \pm 0.004}$ & $\mathbf{0.748 \pm 0.004}$ & $\mathbf{0.726 \pm 0.006}$ & $\mathbf{0.736 \pm 0.009}$ \\
    \midrule
    \multirow{4}{*}{First come - First serve} 
& Gaussian & $0.695 \pm 0.001$ & $0.774 \pm 0.001$ & $0.795 \pm 0.001$ & $0.888 \pm 0.001$ & $0.945 \pm 0.001$ \\ 
& Tobit & $0.693 \pm 0.004$ & $0.772 \pm 0.010$ & $0.750 \pm 0.003$ & $0.785 \pm 0.010$ & $0.790 \pm 0.016$\\ 
& QR & $0.577 \pm 0.007$ & $0.697 \pm 0.005$ & $0.734 \pm 0.003$ & $0.841 \pm 0.011$ & $0.894 \pm 0.007$ \\ 
& Censored QR & $\mathbf{0.574 \pm 0.004}$ & $\mathbf{0.690 \pm 0.005}$ & $\mathbf{0.709 \pm 0.005}$ & $\mathbf{0.774 \pm 0.005}$ & $\mathbf{0.765 \pm 0.004}$ \\
    \bottomrule
    \end{tabular}
    }
    \caption{Comparison of tilted loss summed for all nodes across different models and EV penetration rates. Lower values indicate better performance, and the best-performing model is highlighted in bold.}
\label{tab:tilte_loss_test}
\end{table}

\begin{figure}
\includegraphics[width=0.495\textwidth]{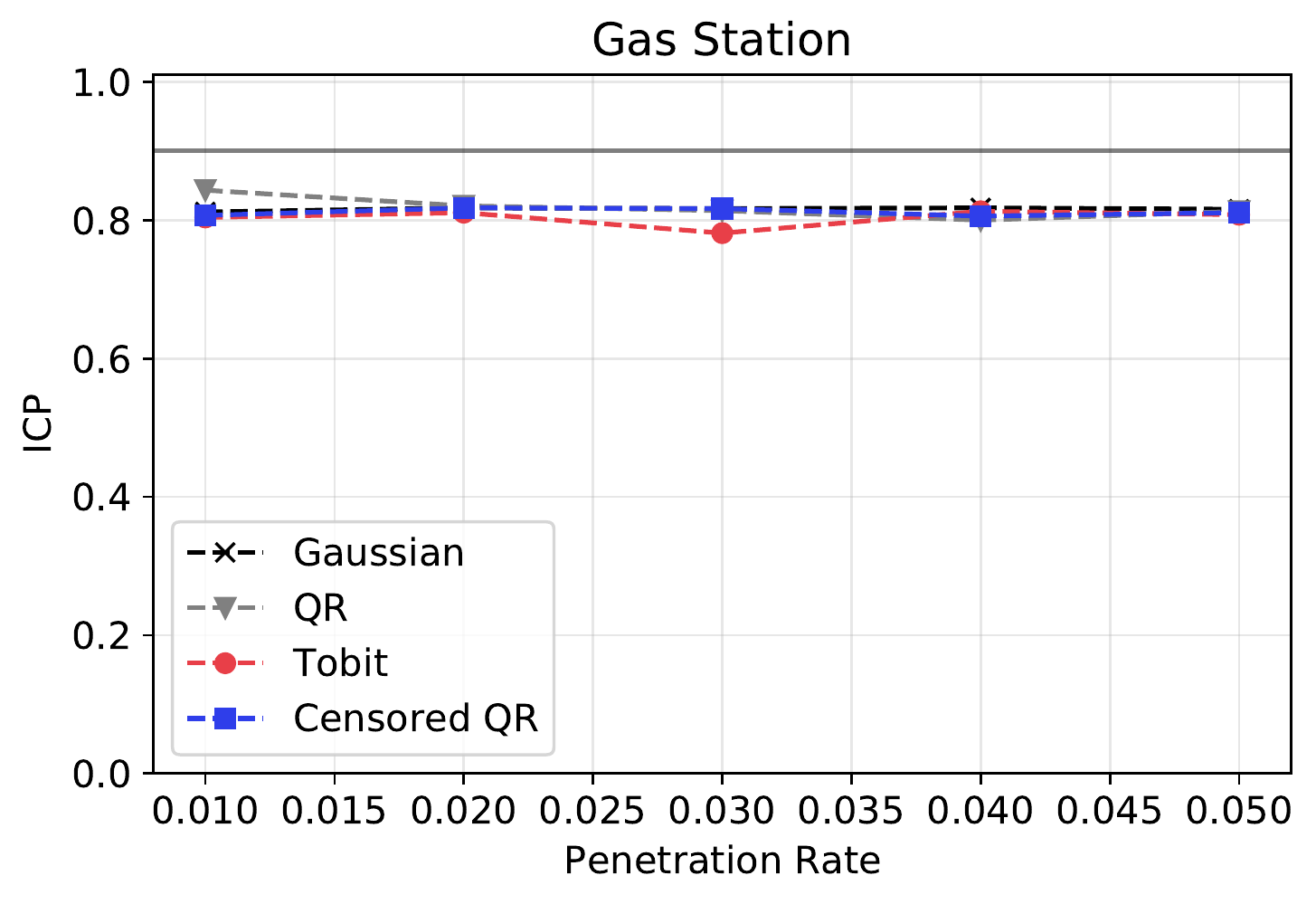}
\hfill
\includegraphics[width=0.495\textwidth]{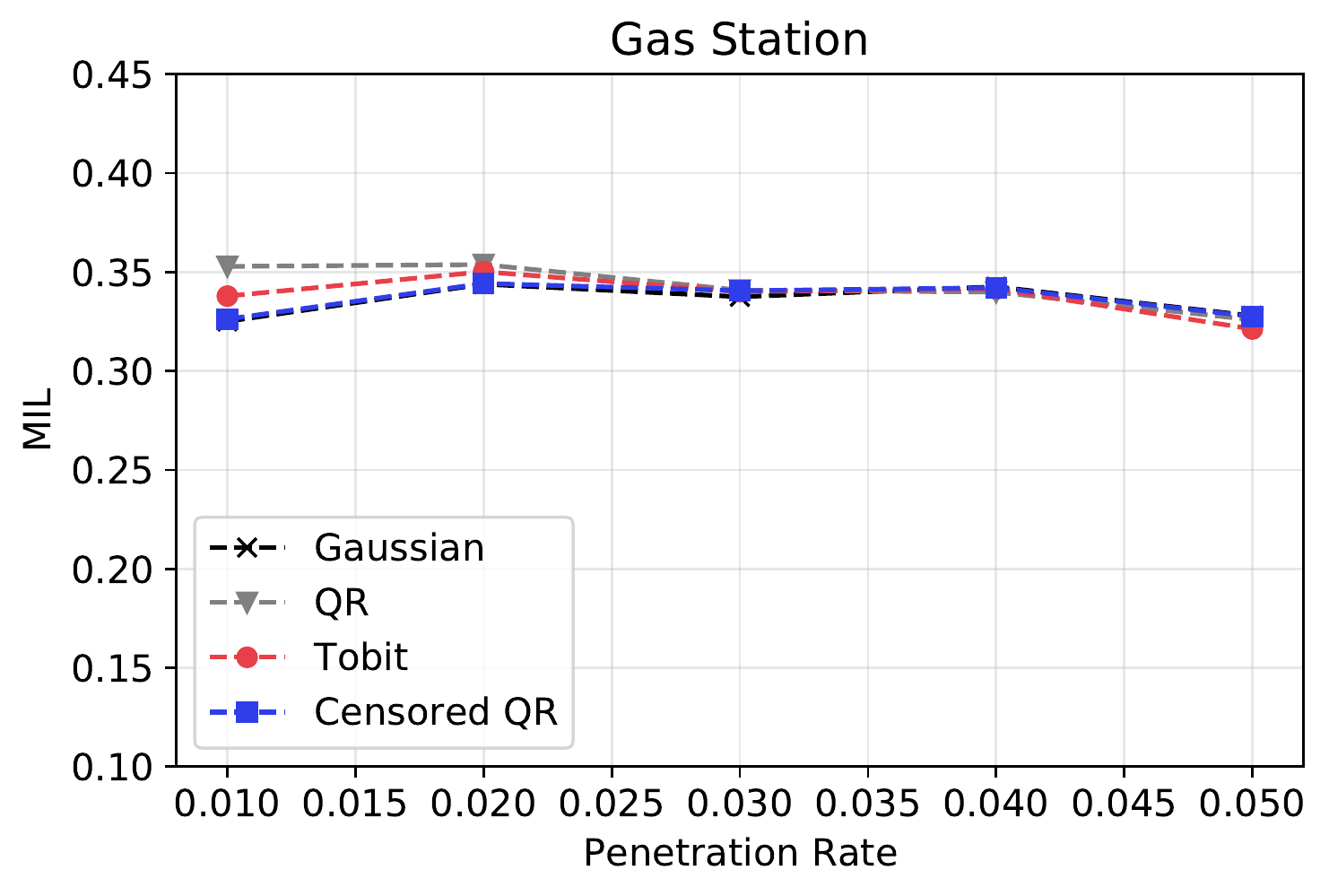}
\includegraphics[width=0.495\textwidth]{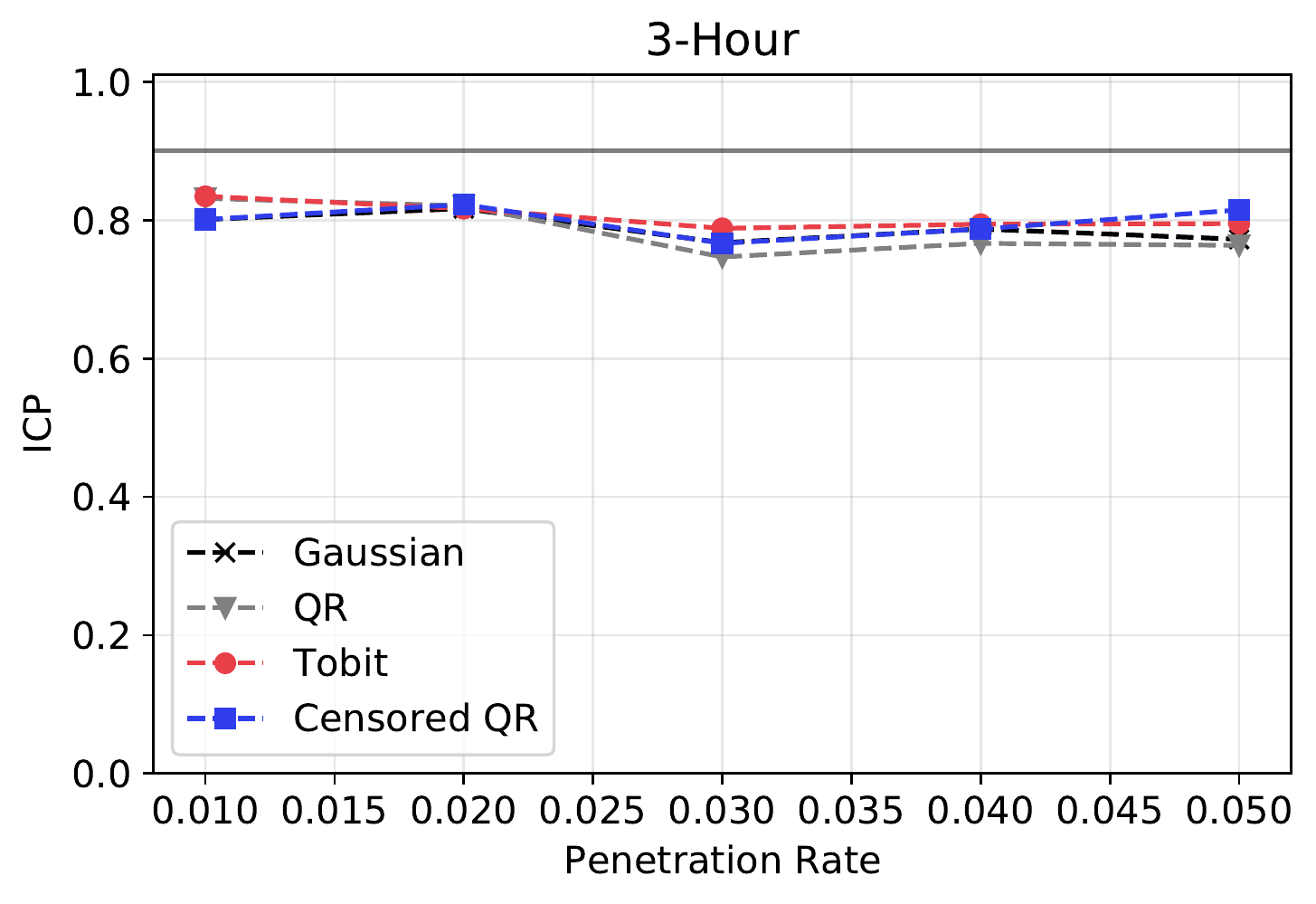}
\includegraphics[width=0.495\textwidth]{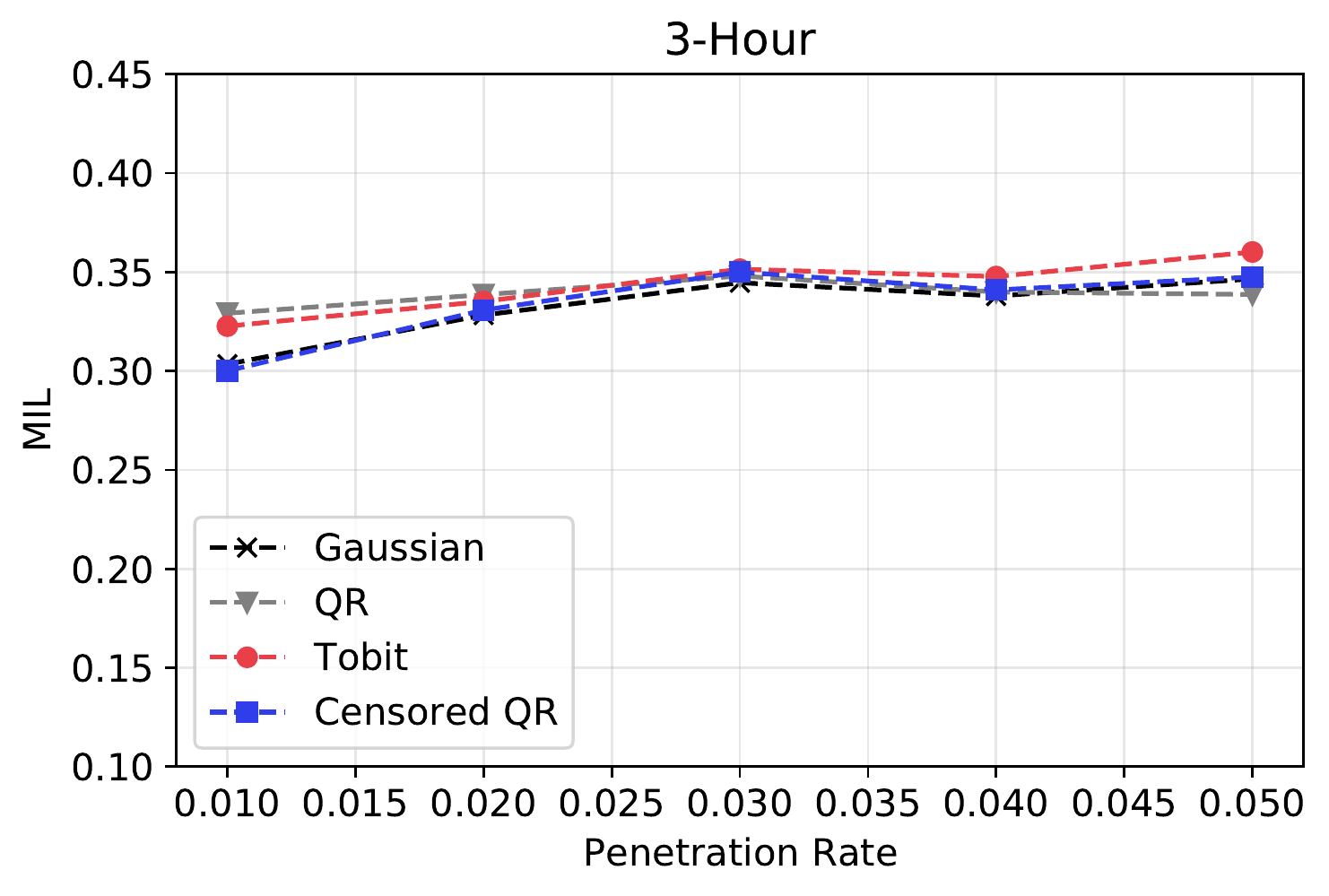}
\includegraphics[width=0.495\textwidth]{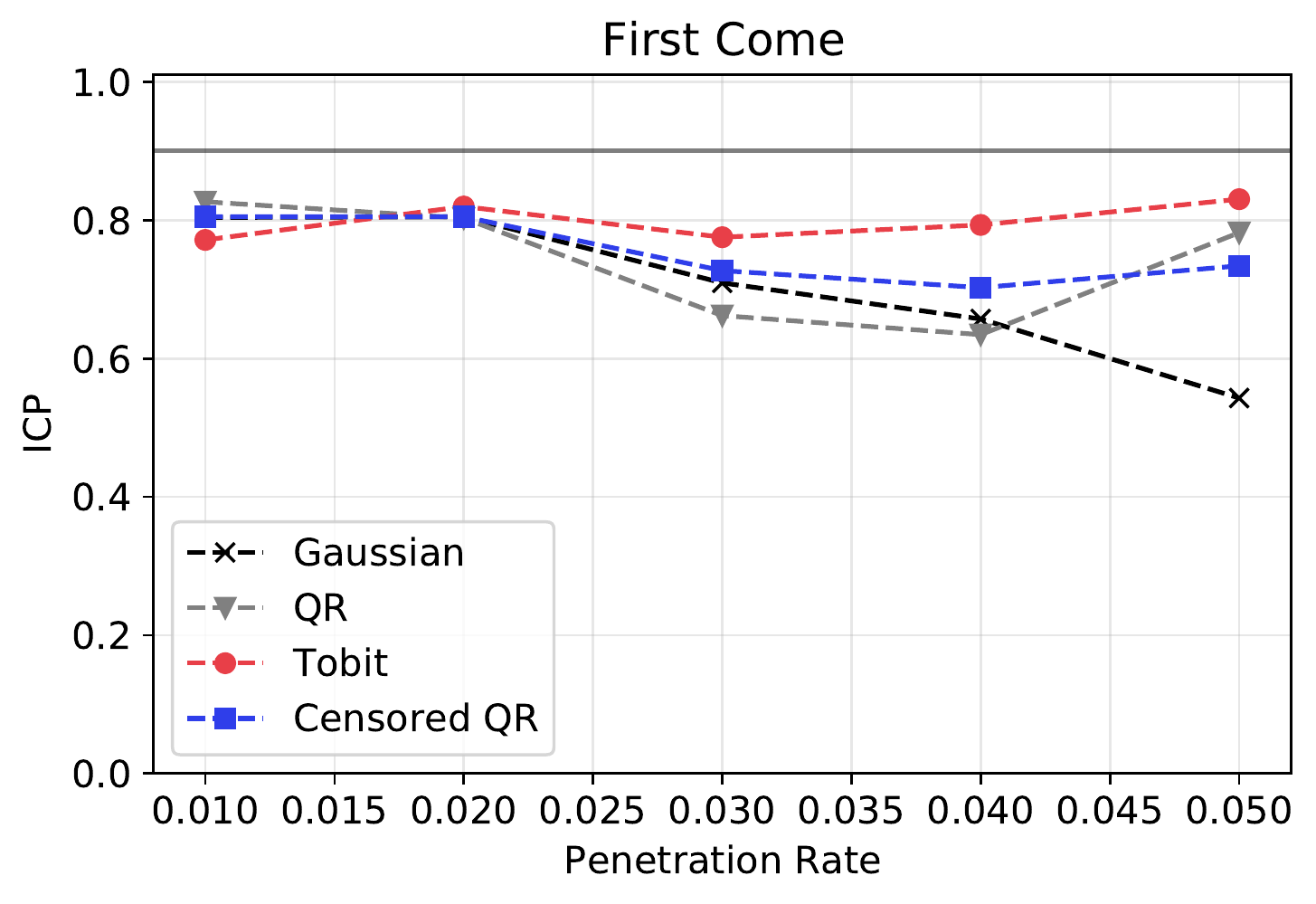}
\includegraphics[width=0.495\textwidth]{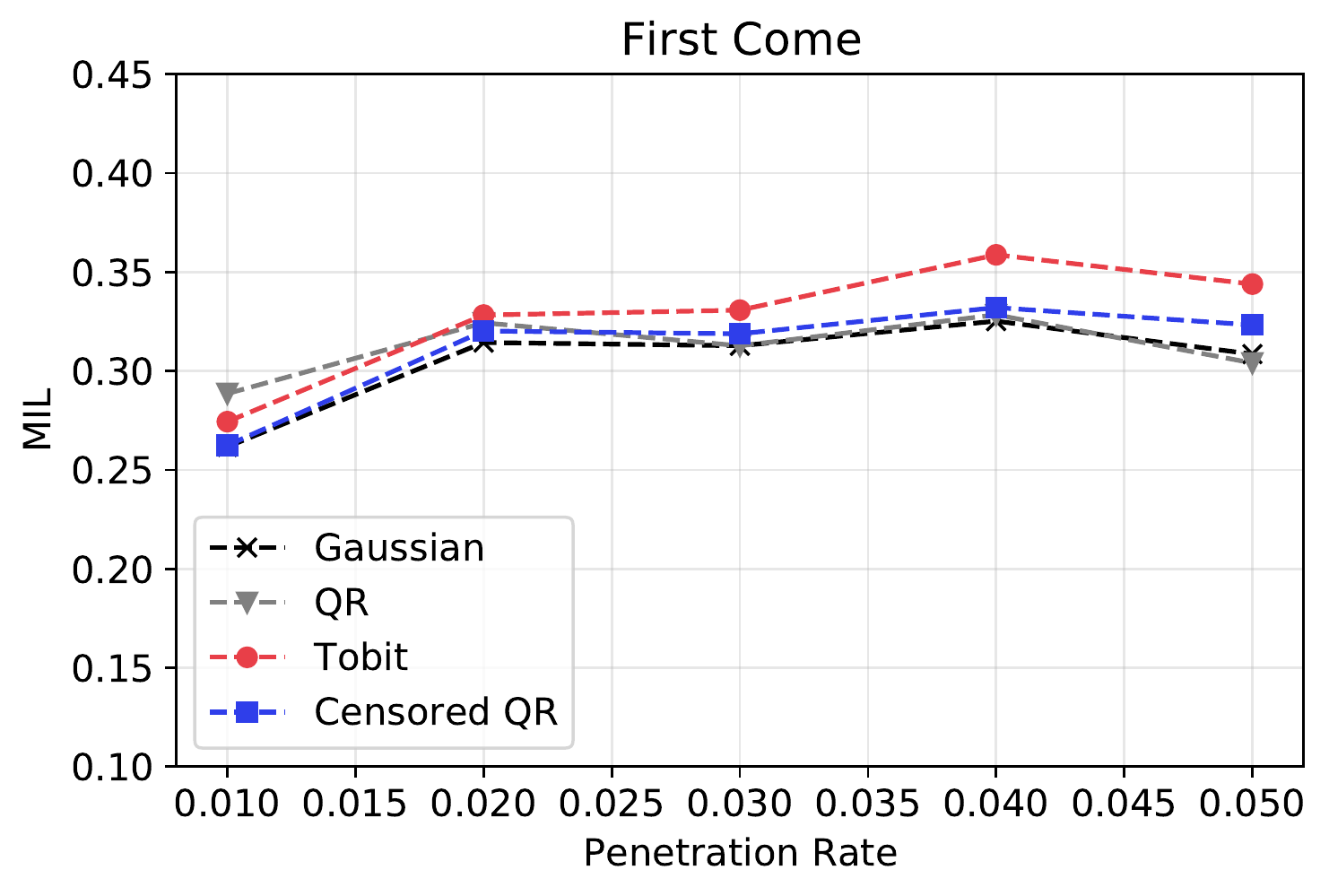}
\caption{The ICP (left) and MIL (right) across the different penetration rates and Queues. In general, the ICP should be close to 0.9 with a lower MIL value, which means that the uncertainty can capture the target distribution while allowing for uncertainty in the predictions.}
\label{fig:uncer}
\end{figure}

In \autoref{fig:uncer}, we present the results of the Interval Coverage Percentage (ICP) and Mean Interval Length (MIL) metrics for the models on different queues and penetration rates. The ICP represents the proportion of observations within the prediction interval, while the MIL measures the average length of the prediction intervals. Since the level of censoring varies between nodes, we present the ICP for the cluster with the highest censorship level.
The Gaussian models, namely Gaussian and Tobit, generally show higher ICP and MIL values, indicating larger prediction intervals than the non-parametric models. As a result, they achieve a higher ICP than the QR models. On the other hand, the QR models exhibit a narrower confidence interval, resulting in lower MIL values, while their ICP is comparable to that of the Gaussian models. Therefore, the best model is the one with an ICP close to $90\%$ and the shortest possible MIL.

\subsection{Application of Censored Demand Modelling}
\begin{figure}[b]
    \centering
    \includegraphics[width=\textwidth]{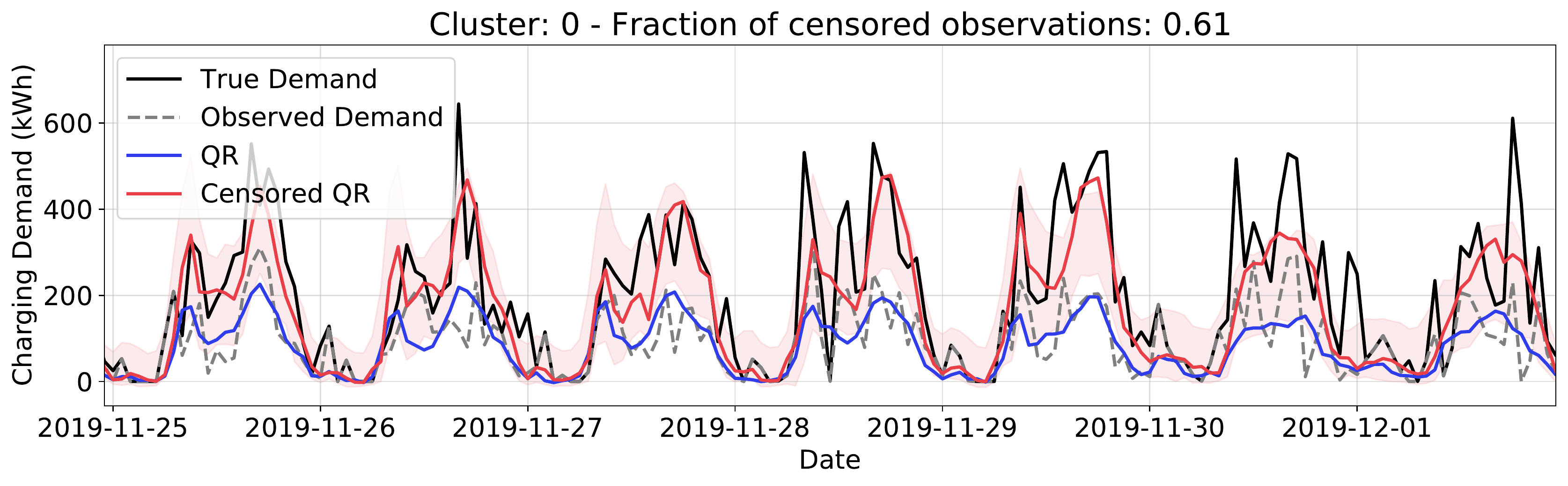}
    \caption{Demand predictions for cluster 0 in the first-come queue scenario, with a 5\% penetration rate. The black line represents the true demand, while the observed demand is shown in grey. The blue and red lines correspond to the predictions obtained using the quantile and censored quantile regression models, respectively. The shaded red prediction interval is defined by the quantiles $q_{\theta,0.05}(x_i)$ and $q_{\theta,0.95}(x_i)$}
    \label{fig:charging_stations_demand_predictions}
\end{figure}

In this section, we discuss the advantages of using the censorship-aware model over the censorship-unaware model and illustrate how it can influence future expansion plans. While infrastructure expansion planning is a complex process, we focus on the benefits of using censorship-aware models.

Cluster 0 is a cluster where a large fraction of observations is censored. The censored QR model better fits the actual true demand than the censorship-unaware QR model (\autoref{fig:charging_stations_demand_predictions}). The QR model provides a relatively accurate fit of the observed demand and tends to forecast the demand sufficiently by forecasting the general temporal patterns. However, the censored QR model provides higher predictions of the charging demand than the system observes. These elevated predictions suggest that potentially expanding the infrastructure in cluster 0 is viable.
One benefit of the censorship-aware models is their estimation of the capacity increase required to meet the true demand. The gap between the predicted demand and observed censored demand could already serve as an indicator for expansion. However, the regressed value and their corresponding uncertainty provide a size of the additions needed to meet all demands. Therefore, censorship-aware models estimate the capacity to meet demand, providing a reasonable basis for data-driven decisions. Additionally, flexible supply could serve peak demand hours due to the variation in demand throughout the day.

\subsection{Competing Services}
We now investigate a scenario where multiple EV charging providers operate in a competitive environment. Our goal is to forecast the true demand for operational expansions from the perspective of a single charging service provider who only observes a fraction (market share) of the total charging demand in the city. We vary the market share from $10\%$ to $95\%$, where a $10\%$ market share means that $90\%$ of the true demand remains unobserved. We do not assume any particular number of competitors in the market and only use data one provider observes. To cluster the chargers, we follow the same method as previously described. 
We sample the market share of chargers in the entire city and censor the demand from all other charging stations.

We report our experiments' results regarding tilted loss, ICP, and MIL in \autoref{tab:comp_test}. We find that the censorship-aware models outperform their unaware counterparts. Specifically, the censored quantile regression model tends to provide a better fit for the charging demand for low market shares while still having a low performance of the ICP and MIL. As the market share increases and the censorship decreases, the performance gap between the censorship-aware and unaware models diminishes. However, for a market share of $95\%$, the censorship-aware models still provide tighter uncertainty in their prediction intervals while still having a high percentage of observations within their intervals.

These results demonstrate the benefits of using censorship-aware models in a competitive environment. The models provide a reliable basis for the charging station's operations and estimates of the demand that competitors can serve. Estimating demand lost to competitors provides a competitive edge for future expansions and the operation of the infrastructure from the perspective of a charging station operator. This estimate can be used for strategic moves \citep{chaojie2021data}.

\begin{table}[bt]
    \centering
    \resizebox{0.95\textwidth}{!}{
    \begin{tabular}{ll|rrrrr}
    \toprule
     & & \multicolumn{5}{c}{Market share} \\
    \thead{Metric} &\thead{Model} & \thead{$10\%$}& \thead{$25\%$} & \thead{$50\%$}  & \thead{$75\%$} & \thead{$95\%$} \\
    \midrule
    \multirow{4}{*}{TL} 
& Gaussian & $3.289 \pm 0.005$ & $1.991 \pm 0.007$ & $1.149 \pm 0.016$ & $0.868 \pm 0.006$ & $0.764 \pm 0.007$  \\ 
& Tobit & $2.799 \pm 0.053$ & $1.667 \pm 0.030$ & $1.027 \pm 0.010$ & $0.838 \pm 0.009$ & $0.765 \pm 0.006$ \\
& QR & $3.120 \pm 0.023$ & $1.779 \pm 0.013$ & $1.019 \pm 0.009$ & $0.775 \pm 0.004$ & $0.686 \pm 0.004$ \\ 
& Censored QR & $\mathbf{2.636 \pm 0.026}$ & $\mathbf{1.538 \pm 0.009}$ & $\mathbf{0.940 \pm 0.006}$ & $\mathbf{0.754 \pm 0.007}$ & $\mathbf{0.684 \pm 0.006}$ \\
    \midrule
    \multirow{4}{*}{ICP} 
& Gaussian & $0.566 \pm 0.002$ & $0.660 \pm 0.005$ & $0.766 \pm 0.009$ & $0.820 \pm 0.004$ & $0.867 \pm 0.004$ \\ 
& Tobit & $0.567 \pm 0.011$ & $0.713 \pm 0.016$ & $0.807 \pm 0.014$ & $0.839 \pm 0.008$ & $0.866 \pm 0.005$ \\
& QR & $0.398 \pm 0.008$ & $0.557 \pm 0.033$ & $0.725 \pm 0.011$ & $0.772 \pm 0.011$ & $0.812 \pm 0.015$ \\
& Censored QR & $\mathbf{0.389 \pm 0.016}$ & $\mathbf{0.642 \pm 0.014}$ & $\mathbf{0.732 \pm 0.023}$ & $\mathbf{0.768 \pm 0.026}$ & $\mathbf{0.809 \pm 0.008}$ \\
    \midrule
    \multirow{4}{*}{MIL} 
& Gaussian & $0.225 \pm 0.006$ & $0.295 \pm 0.003$ & $0.312 \pm 0.006$ & $0.315 \pm 0.005$ & $0.331 \pm 0.004$ \\ 
& Tobit & $0.246 \pm 0.009$ & $0.336 \pm 0.011$ & $0.349 \pm 0.005$ & $0.333 \pm 0.007$ & $0.332 \pm 0.006$ \\ 
& QR & $0.164 \pm 0.005$ & $0.263 \pm 0.003$ & $0.284 \pm 0.006$ & $0.296 \pm 0.006$ & $0.318 \pm 0.005$ \\
& Censored QR & $\mathbf{0.256 \pm 0.005}$ & $\mathbf{0.323 \pm 0.005}$ & $\mathbf{0.319 \pm 0.003}$ & $\mathbf{0.313 \pm 0.004}$ & $\mathbf{0.323 \pm 0.003}$ \\
    \bottomrule
    \end{tabular}
    }
    \caption{Model performance for increasing market share with three different evaluation metrics, Tilted loss (TL), Interval Coverage Percentage (IPC), and Mean Interval Length (MIL). The bold values indicate the model with the lowest Tilted loss}
    \label{tab:comp_test}
\vspace{-5mm}
\end{table}

\section{Discussion}
\label{sec:discussion}
Throughout our analysis, experiments and results, we have argued that the observed EV charging demand from charging records is censored and, thus, does not reflect the actual demand for charging. 
We conducted a counterfactual study using ICE GPS trajectories and battery models to investigate this hypothesis. 

One advantage of our counterfactual study is that it employs real-world censoring instead of artificial censoring schemes commonly used in censored modelling \citep{huttel2022modeling, gammeli2020estimating}.
The censoring is occurring due to the inclusion of queues and limited supply. Therefore, the censoring scheme consistent with actual charging infrastructure provides a more accurate representation of how the censoring occurs.
Another advantage of using the ICE GPS trajectories is that early EV adopters do not influence them. 
Hence, they represent the desired behaviour of users when they are not constrained by needing to recharge their cars. 
Our counterfactual study provides an initial indication of the degree of underestimation of the observed demand that is caused by censorship.
Another advantage of using the proposed method to estimate the censored demand is that it allows for the inference of latent demand from observational data.
In this work, the counterfactual study is \emph{just} used to validate our censored models with a realistic censorship scheme compared to an artificial one.

While ICE GPS trajectories were used in our study, naturally, EV trajectories would provide a better proxy for the charging demand, as behavioural differences exist between ICE and EV users. 
For example, the ICE cars do not need to recharge; hence, they represent the desired behaviour of users as they are not constrained by needing to recharge their cars. On the other hand, EV users may be more conservative due to the slow charging speed and the shortage of charging facilities, which could result in shorter trips per charge.
This aspect, however, is also likely to change with more efficient and widespread technology.

The counterfactual study conducted in this work shares many similarities with agent-based simulations but with the distinction that individual travel behaviours, such as origins, destinations, and parking times, are determined by GPS trajectories. The counterfactual nature of the study represents the "what-if" scenario of running the same trips with different assumptions of the EV. However, validating the simulation is challenging since data from the period is not obtainable.
Future studies could incorporate additional factors, such as departure time and route, to improve the simulation's accuracy. Collecting more comprehensive data on EV charging behaviour would also be beneficial to improve the results' reliability and inform the development of appropriate charging infrastructure and policies.

Our study acknowledges that our simplistic charging and battery models may underestimate the charging demand, as we do not account for other key travel metrics such as speed, acceleration, and ambient temperature \citep{mcnerney2017tripenergy}. These factors also affect battery consumption and result in variations in the charging demand for each trip. Therefore, the censored charging demand may be greater than we present here, highlighting the need for censorship-aware models to model EV charging demand accurately. While a more behaviourally sophisticated approach, such as a discrete choice model, would be ideal, it requires data currently unavailable, such as the social demographics and preferences of the users.

\section{Conclusion and future research directions}
\label{sec:conclusion}
In conclusion, we have addressed the issue of censorship in modelling electric vehicle charging demand by extending machine learning models to be censorship-aware.
We argue that the demand observed by charging records is censored due to lost opportunities to competition or supply limitations, which can negatively impact censorship-unaware models' ability to model the demand. It is essential to address the gap between lost and observed demand to facilitate efficient and cost-effective expansions and operations of charging stations.

We propose several methods to account for censorship in modelling EV charging demand and extend them to spatiotemporal problems, namely the Tobit model and the censored quantile regression. 
An advantage of the proposed method is that it can be applied to any observed charging records. 
Specifically, we used a set of GPS trajectories from ICE cars in Copenhagen, Denmark. We studied how they would have behaved by modelling their behaviour and charging dynamics, assuming the cars were EVs.
This leads to various amounts of censorship across space and time.
Using the observed charging demand from this study, we demonstrate the superiority of censorship-aware models over censorship-unaware models in accurately modelling the latent distribution of charging demand.
The experiments show that the censored quantile regression model obtains a better fit and uncertainty estimation of the demand than the Tobit model. Furthermore, We compared censorship-aware and unaware models in a competitive environment with different market share amounts. We found that censorship-aware models are superior in modelling the demand, which could provide an edge in a competitive environment. 

Furthermore, while this study focuses on two specific scenarios of demand censorship (frustrated demand due to occupied charging stations; diminished demand due to competing services), other types of latent, lost, or unobserved demand exist. This includes exploring additional factors such as charging station unavailability, faulty equipment, pricing structures, and user behaviour patterns that can contribute to censored demand. Future research should also investigate the implications of incorporating mobile charging stations into the existing infrastructure, as they can be used to uncover the true demand.


\section{Acknowledgements}
The research leading to these results has received funding from the Independent Research Fund Denmark (Danmarks Frie Forskningsfond) under grant no. 0217-00065B.

\newpage
\bibliographystyle{apalike}
\bibliography{ref}

\newpage 

\appendix
\section{Counterfactual study - appendix}
\label{sec:counterfactualstudy}
\subsection{Data}
\begin{figure}[b]
    \centering
    \includegraphics[width=0.95\textwidth]{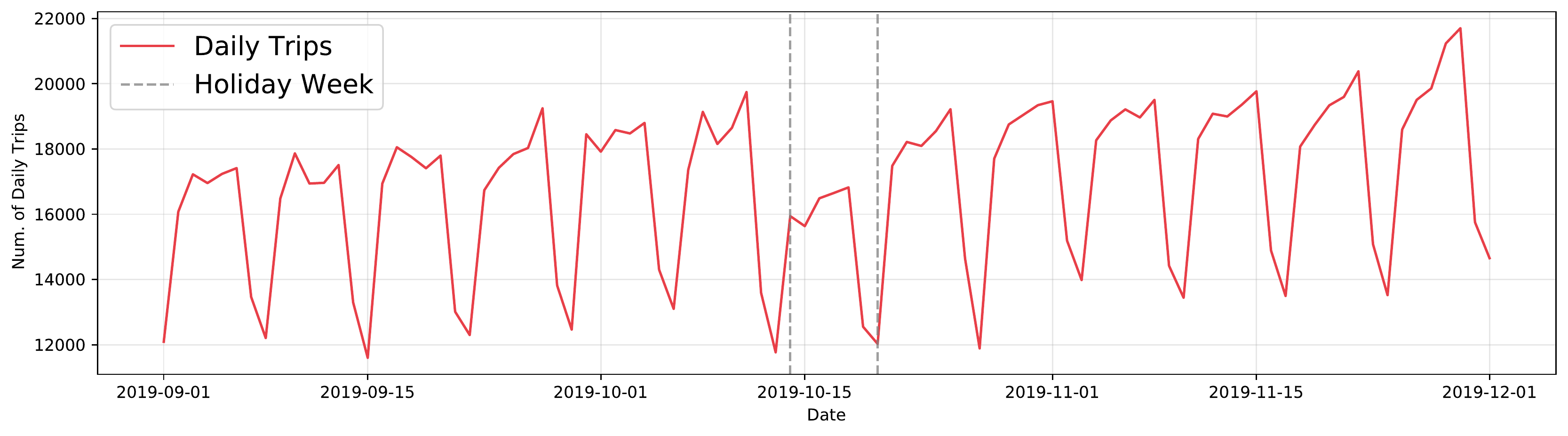}
    \caption{Overview of the total daily trips in the data set. From the 14th to the 20th of October, there is a drop in daily trips due to a holiday week in Denmark. Therefore, there are fewer trips due to a reduced traffic flow from commuting.}
    \label{fig:num_daily_trips_app}
\end{figure}

As mentioned, the data consist of GPS trajectories from ICE cars in Denmark, where locations are saved every 20 seconds. 
Each GPS location contains an ID of the car and can be stitched together into a complete GPS trajectory, which forms a trip for the car.
Each trip contains a start coordinate, an end coordinate, and the distance driven by a car.
Cars' parking time can be inferred as the time between trips.
To ensure the users' privacy, the trajectories are randomised by adding a noise distance between 50 and 500 meters to each trip's endpoints. 
The GPS trajectories of the cars were collected over three months, from September to November 2019 and are uniformly distributed across the country (Denmark) and vehicle segments. 
The GPS trajectories do not suffer from the biases identified in other EV studies, namely, having early adopters bias the behaviour in the data sets.
In total, 32664 cars are observed in the capital region (Copenhagen), which accounts for $5.71\%$ of the cars in the city (A total of 571627 cars for the period \citep{statbank2022}). 
In the observations period, the  penetration rate of EVs in Copenhagen is $1\%$, and for the fall of 2022, the penetration rate has increased to $2.5\%$~\citep{statbank2022}. We visualise the number of daily trips for the entire period in \autoref{fig:num_daily_trips_app}. 
In total, we have 1550509 trips with a median of 10 trips pr. car.

\paragraph{Charging Stations in Copenhagen}
To model the charging infrastructure in Copenhagen, we use the charging infrastructure from 2021. We scraped the infrastructure from \citet{uppladdning_2021}, an open-source map containing locations of EV charging stations. The scraped chargers have different charging power ranging from slow chargers with 3.7 kW charging to fast chargers with 150 kW.
These provide us with the locations and the power for each charging station in the city, which we use to determine where the ICE cars would have charged, assuming they were EVs.

\paragraph{Market shares of Electric vehicles}
\label{sec:marksetshareev}
Before we describe our battery and charging models, we generate the fleet of EVs based on the distribution of EVs in Denmark.
To generate the EV fleet, we sample the EV characteristics for each car according to their market share distribution.
We report the ten most popular EV models in Denmark can be seen in \autoref{tab:ev_info}. 

\begin{table}[!h]
    \centering
    \begin{tabular}{llrrrr}
    \toprule 
    \thead{Ranking} &\thead{Model} &  \thead{Count} &  \thead{Range (km)} &    \thead{Battery capacity (kWh)}  \\  \midrule
    1&Tesla Model 3 SR      &   8183 &  380.0 &                    57.0   \\
    2&Renault Zoe        &   4050 &  315.0  &                    52.0  \\
    3&Tesla Model S      &   3915 &  560.0  &                    95.0  \\
    4&Volkswagen ID.3 EV &   3353 &  350.0  &                    58.0  \\
    5&Nissan Leaf        &   3033 &  225.0  &                    37.0  \\
    6&Hunydai Kona BEV   &   2948 &  395.0  &                    64.0  \\
    7&Volkswagen ID.4 EV &   2473 &  400.0  &                    77.0  \\
    8&Kira Niro EV       &   1890 &  370.0  &                    64.0 \\
    9&BMW i3             &   1642 &  235.0  &                    37.9 \\
    10&Volkswagen e-Up!   &   1370 &  205.0  &                    32.3 \\
    -&Others             &  17399 &  313.0  &                    60.0 \\
    \bottomrule
    \end{tabular}
    \caption{Top 10 most popular EV models in Denmark \citep{fleet2022} with manufacturing standards from \citep{evdatabase2020}.}
    \label{tab:ev_info}
\end{table}

\paragraph{Generation of EV fleet}
We generate a fleet of EVs from the ICE using the market shares. Where each EV corresponds to an ICE, with the same trips, for each EV we keep track of the \emph{SoC}, \emph{Range} and \emph{battery capacity} for the vehicle. 

\subsection{Battery Model}
Firstly, we turn to our modelling of the Lithium ION batteries in the EV fleet.
Initially, we set the SoC levels of the fleet to follow a truncated normal distribution as $SoC \sim \mathcal{N}(0.6,0.2), 0.20 \leq SoC \leq 1$.
We assume there is a linear relationship between the distance of a trip and State of charge consumption for a trip and model the SoC consumption (or the decrease in the SoC levels) as:

\begin{equation}
\label{eq:cons}
    \text{Consumption pr. trip} = \frac{\text{Distance of trip}}{\text{Range of car type}}\, .
\end{equation}

After one trip, we update the EV with the SoC for the next trip as:
\begin{equation}
\label{eq:soc}
\text{SoC}_{t+1} =   \text{SoC}_{t}-\text{Consumption pr. trip}\, .
\end{equation}

\subsubsection{Charging decision}
We model the charging decision with a smooth beta function. We evaluate the probability of charging after a trip as a CDF of a beta distribution with the initial $x_i$ and the final SoC $x_f$ levels. \citep{hipolito2022charging}
The PDF of a beta distribution is
\begin{equation}
    f_\beta(a, b ; x)=\frac{x^{a-1}(1-x)^{b-1}}{\Gamma(a) \Gamma(b) / \Gamma(a+b)} \, .
\end{equation}
We model the probability of charging with a $SoC \leq x$ as the CDF ($F_\beta(a, b ; x)$)  of the survival function
\begin{equation}
S_\beta(a, b ; x)=1-F_\beta(a, b ; x)=1-\int_0^x f_\beta(a, b ; t) d t
\end{equation}
Using this equation for the decision to charge can lead to counter intuitive examples of charging, where cars charge even at high levels of $SoC$.
Therefore we make the vehicles context-aware by assessing the willingness to charge based on the initial $SoC$ ($x_i$) and the final $SoC$ ($x_f$) of a trip.
We compute the willingness to charge based on the differences between the CDFs for the initial SoC and the final SoC, normalised to the probability of charging at the start÷

\begin{equation}
\label{eq:willignes}
W_\beta\left(a, b ; x_f, x_i\right)=\left[F_\beta\left(a, b ; x_i\right)-F_\beta\left(a, b ; x_f\right)\right] / F_\beta\left(a, b ; x_i\right) .
\end{equation}

Using the willingness to charge, we random draw if a car decides to charge. The parameters $a$ and $b$ are estimated via non-linear regression
We use default parameters for the distributions, and the reader is referred to \citep{hipolito2022charging} for a larger discussion on the willingness to charge and the estimation of $a$ and $b$.

\subsection{Where to charge}
The willingness to charge gives us a willingness score between 0 and 1. We randomly draw if you charge or not based on the willingness score. 
We employ a simple discrete choice model to decide where to charge if the car needs to charge 
We identify the five closest chargers based on the trip end point and compute their utility as the distance between the endpoint of the trip and the charger location. For each station of the five closest, we draw the probability:
\begin{equation}
\label{eq:Wheretocharger}
    P_{j} = \frac{\exp (D_j)}{\sum\exp(D_j)} \, ,
\end{equation}

where $D_j$ is the distance between the charging station and the end trip point the trip and $P_j$ is the probability of going to station $j$. We then randomly select a charging station from the $P_j$ of the five closest charging stations.

\subsubsection{Charging model}
We model the decision to charge as a probabilistic choice between the 
Once a charger has been chosen, we approximate the charging of the EV with a piece-wise linear model \citep{marjan2020optimal}. The time spent charging will be dependent on the queue for the study. Firstly, we compute the time it takes for the car to have an SoC of more than $80\%$ as:

\begin{equation}
T_{80} = 0.8\left(\frac{\text{Battery capacity}-SoC\cdot \text{Battery capacity}}{\text{Charger power}}\right) \, .
\end{equation}

\begin{figure}[tb]
    \centering
    \includegraphics[width=0.85\textwidth]{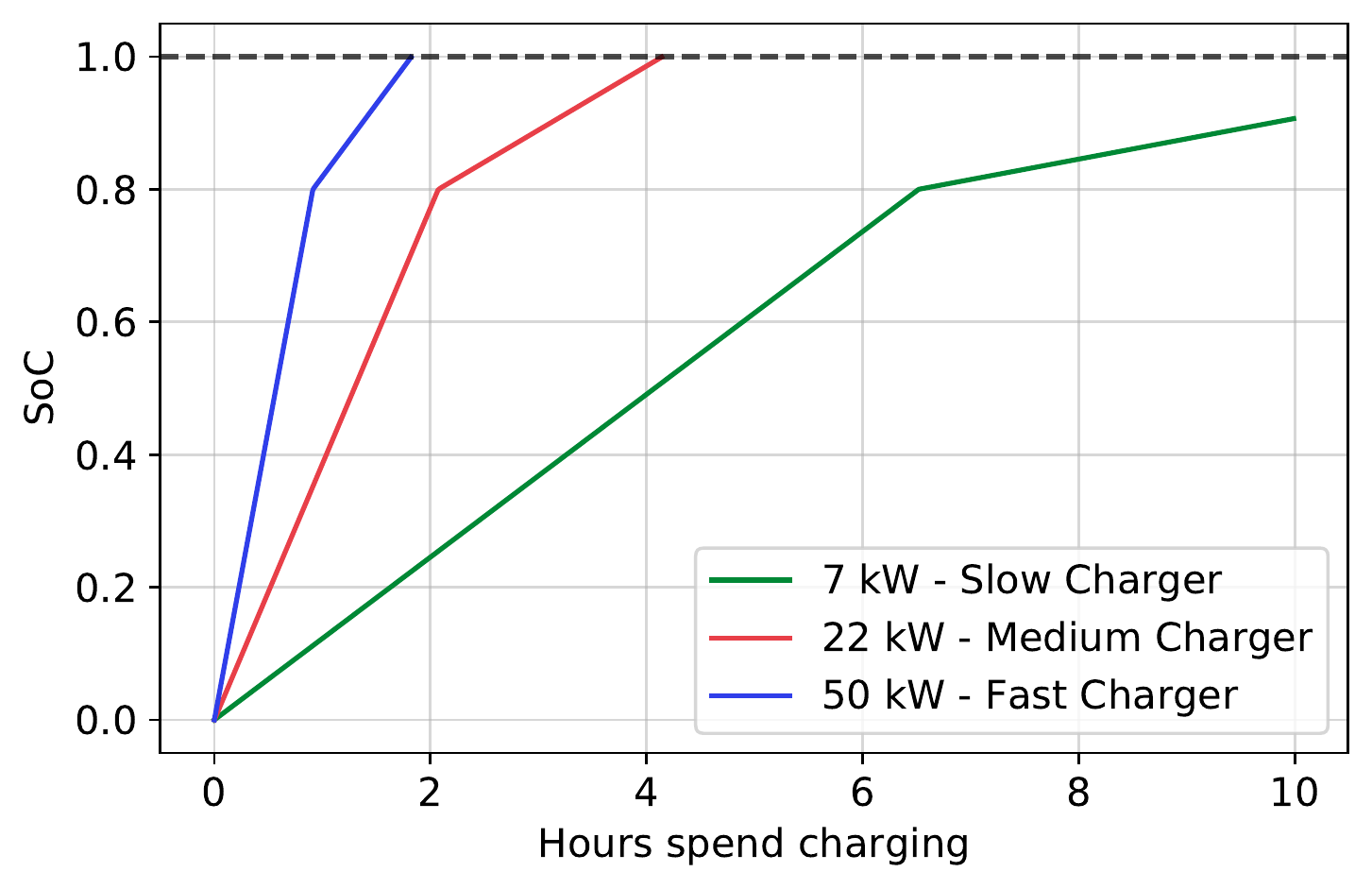}
    \caption{Charging curves for a \emph{Tesla Model 3} with a battery capacity of 57 kWh and three different charging capacities. A 7kW slow charger, a 22 kW medium charger and a 50 kW fast charger. }
    \label{fig:tesla_charging}
\end{figure}

We then compute the new SoC for the car based on the following. In \autoref{fig:tesla_charging}, we show an example of the charging equation for three different charger types:
\begin{equation}
    \text{SoC}_{t+1} = \begin{cases}(\text{SoC}_{t} + \text{charging time}*\frac{\text{Charger power}}{\text{Battery capacity}})  & \text{if charging time} <  T_{80} \\
  0.8+0.25((\text{charging time above } T_{80})*\frac{\text{Charger power}}{\text{Battery capacity}}) & \text{else}
  \end{cases}\,.
\label{eq:energyreq}
\end{equation}

Finally, we store the energy required for the charging. If a vehicle did not charge due to occupied charging stations, we assume it would have charged $80\%$ SoC and stored this as censored demand.

\subsection{Summary of the counterfactual study}
Using the points made above, we conduct the counterfactual study. We vary the number of sample EVs to match the different penetration rates. In 2019, the penetration rate of EVs was roughly 1\% and in 2022, is roughly 3\%. The GPS trajectories themselves have a penetration rate of 5\%. We run the study with three different queuing models. We sample the car distribution according to the count distribution in \autoref{tab:ev_info}. Each car follows the procedure in Algorithm \ref{alg:overview}.

\begin{algorithm}
\caption{Overview of the counterfactual study.}\label{alg:overview_appendix}
\begin{algorithmic}
\State Generate a fleet of electric vehicles.
\State Set queuing model.
\State Set SoC for the fleet of cars $SoC \sim \mathcal{N}(0.6,0.2), 0.20 \leq SoC \leq 1$.
\For{each GPS trajectory}
        \State Find the car id of the GPS trajectory.
        \State Compute energy consumption for the car (\autoref{eq:cons}).
        \State Update SoC-level for the car (\autoref{eq:soc}).
        \State Compute willingness to charge for the car (\autoref{eq:willignes}).
        \State Compute were to charge based on the end location of the trip (\autoref{eq:Wheretocharger}).
        \State Add the car to the queue.
        \State Compute energy demand based on the charging time (\autoref{eq:energyreq}).
        \State Save the energy demand from the charging event.
\EndFor
\end{algorithmic}
\end{algorithm}

\newpage
\section{Censoring across the different clusters}
\label{sec:censoring_across}

\begin{table}[h]
    \centering
    \begin{tabular}{ll|rrrrr}
    \toprule
    & & \multicolumn{5}{c}{Penetration Rate} \\
    \thead{Queue} & \thead{Cluster} & \thead{1.0\%}& \thead{2.0\%} & \thead{3.0\%}  & \thead{4.0\%} & \thead{5.0\%} \\
    \midrule
    \multirow{9}{*}{Gas station} 
&0 & $0.22\%$ &  $1.26\%$ &  $2.80\%$ &  $5.08\%$ &  $8.11\%$\\ 
&1 & $0.04\%$ &  $0.09\%$ &  $0.22\%$ &  $0.49\%$ &  $1.09\%$\\ 
&2 & $0.00\%$ &  $0.00\%$ &  $0.00\%$ &  $0.18\%$ &  $0.36\%$\\ 
&3 & $0.00\%$ &  $0.05\%$ &  $0.27\%$ &  $0.22\%$ &  $0.63\%$\\ 
&4 & $0.0v\%$ &  $0.05\%$ &  $0.14\%$ &  $0.50\%$ &  $1.40\%$\\ 
&5 & $0.00\%$ &  $0.05\%$ &  $0.18\%$ &  $1.13\%$ &  $2.17\%$\\ 
&6 & $0.00\%$ &  $0.05\%$ &  $0.09\%$ &  $0.09\%$ &  $0.36\%$\\ 
&7 & $0.00\%$ &  $0.40\%$ &  $0.67\%$ &  $1.91\%$ &  $3.09\%$\\ 
&8 & $0.00\%$ &  $0.18\%$ &  $0.31\%$ &  $0.86\%$ &  $2.22\%$\\ 
&9 & $0.04\%$ &  $0.10\%$ &  $0.45\%$ &  $0.49\%$ &  $1.22\%$\\ 
    \midrule
    \multirow{9}{*}{3-hour} 
&0 & $2.08\%$ &  $9.74\%$ &  $20.80\%$ &  $31.30\%$ &  $41.96\%$ \\
&1 & $0.27\%$ &  $1.00\%$ &  $4.08\%$ &  $6.25\%$ &  $11.69\%$\\ 
&2 & $0.00\%$ &  $0.22\%$ &  $0.18\%$ &  $0.54\%$ &  $1.45\%$\\ 
&3 & $0.36\%$ &  $1.63\%$ &  $2.72\%$ &  $4.80\%$ &  $9.15\%$\\ 
&4 & $0.09\%$ &  $0.45\%$ &  $1.72\%$ &  $3.27\%$ &  $5.66\%$\\ 
&5 & $0.14\%$ &  $0.50\%$ &  $1.22\%$ &  $3.53\%$ &  $5.89\%$\\ 
&6 & $0.04\%$ &  $0.36\%$ &  $0.58\%$ &  $1.40\%$ &  $2.62\%$\\ 
&7 & $0.27\%$ &  $1.54\%$ &  $3.72\%$ &  $7.52\%$ &  $10.24\%$\\ 
&8 & $0.63\%$ &  $3.27\%$ &  $7.11\%$ &  $10.87\%$ &  $13.18\%$\\ 
&9 & $0.27\%$ &  $1.04\%$ &  $2.22\%$ &  $5.21\%$ &  $8.47\%$\\ 
    \midrule
    \multirow{9}{*}{First come - First serve} 
&0 & $4.17\%$ &  $17.68\%$ &  $38.70\%$ &  $54.64\%$ &  $61.03\%$\\ 
&1 & $0.77\%$ &  $2.36\%$ &  $7.11\%$ &  $12.00\%$ &  $20.29\%$\\ 
&2 & $0.04\%$ &  $0.36\%$ &  $0.58\%$ &  $2.53\%$ &  $3.12\%$\\ 
&3 & $0.86\%$ &  $3.49\%$ &  $6.57\%$ &  $10.74\%$ &  $16.31\%$\\ 
&4 & $0.13\%$ &  $1.99\%$ &  $4.17\%$ &  $6.62\%$ &  $11.28\%$\\ 
&5 & $0.22\%$ &  $1.04\%$ &  $5.71\%$ &  $13.68\%$ &  $24.33\%$\\ 
&6 & $0.22\%$ &  $0.63\%$ &  $1.31\%$ &  $2.45\%$ &  $6.57\%$\\ 
&7 & $0.54\%$ &  $2.62\%$ &  $7.43\%$ &  $12.96\%$ &  $20.16\%$\\ 
&8 & $1.27\%$ &  $5.2\%$ &  $12.96\%$ &  $22.33\%$ &  $31.17\%$\\ 
&9 & $0.59\%$ &  $2.22\%$ &  $4.80\%$ &  $11.69\%$ &  $16.67\%$\\ 
    \bottomrule
    \end{tabular}
    \caption{Percentage of hours where the observed demand is censored across the different queues, penetration rates and clusters.}
\label{tab:ce}
\end{table}

\end{document}